\documentclass[journal]{IEEEtran}
\usepackage{generic}
\usepackage{cite}
\usepackage{amsmath,amssymb,amsfonts}
\usepackage{algorithmic}
\usepackage{graphicx}
\usepackage{textcomp}

\usepackage{caption}
\usepackage{subcaption}
\usepackage[linesnumbered, ruled]{algorithm2e}
\usepackage{booktabs}
\usepackage{amsmath}
\usepackage{amssymb}
\usepackage{multirow}
\usepackage{makecell}

\usepackage{hyperref}

\begin{document}
\title{CDNet: Contrastive Disentangled Network for Fine-Grained Image Categorization of Ocular B-Scan Ultrasound}
\author{Ruilong Dan*, Yunxiang Li*, Yijie Wang, Gangyong Jia, Ruiquan Ge, Juan Ye, Qun Jin, Yaqi Wang$^{\dagger}$
\thanks{The work was supported by the  National Natural Science Foundation of China under Grant No. U20A20386.}
\thanks{* Co-first author. }
\thanks{$\dagger$ Corresponding author.}
\thanks{Ruilong Dan, Yunxiang Li, Gangyong Jia, and Ruiquan Ge are with College of Computer Science and Technology, Hangzhou Dianzi University, Hangzhou, China (e-mail: drldyx20xx@hdu.edu.cn, li1124325213@hdu.edu.cn, gangyong@hdu.edu.cn, and gespring@hdu.edu.cn).}
\thanks{Yijie Wang, Juan Ye are with Department of Ophthalmology, the Second Affiliated Hospital of Zhejiang University, Hangzhou, China (e-mail: wangyj1103@zju.edu.cn, yejuan@zju.edu.cn).}
\thanks{Qun Jin is with the Department of Human Informatics and Cognitive Sciences, Faculty of Human Sciences, Waseda University, Tokyo, Japan (e-mail: jin@waseda.jp).}
\thanks{Yaqi Wang is with the College of Media Engineering, Communication University of Zhejiang, Hangzhou, China (e-mail: wangyaqi@cuz.edu.cn).}}

\maketitle

\begin{abstract}
Precise and rapid categorization of images in the B-scan ultrasound modality is vital for diagnosing ocular diseases. Nevertheless, distinguishing various diseases in ultrasound still challenges experienced ophthalmologists. Thus a novel contrastive disentangled network (CDNet) is developed in this work, aiming to tackle the fine-grained image categorization (FGIC) challenges of ocular abnormalities in ultrasound images, including intraocular tumor (IOT), retinal detachment (RD), posterior scleral staphyloma (PSS), and vitreous hemorrhage (VH). Three essential components of CDNet are the weakly-supervised lesion localization module (WSLL), contrastive multi-zoom (CMZ) strategy, and hyperspherical contrastive disentangled loss (HCD-Loss), respectively. These components facilitate feature disentanglement for fine-grained recognition in both the input and output aspects. The proposed CDNet is validated on our ZJU Ocular Ultrasound Dataset (ZJUOUSD), consisting of 5213 samples. Furthermore, the generalization ability of CDNet is validated on two public and widely-used chest X-ray FGIC benchmarks. Quantitative and qualitative results demonstrate the efficacy of our proposed CDNet, which achieves state-of-the-art performance in the FGIC task. Code is available at: \href{https://github.com/ZeroOneGame/CDNet-for-OUS-FGIC}{https://github.com/ZeroOneGame/CDNet-for-OUS-FGIC}.
\end{abstract}


\section{Introduction}
\label{sec:Introduction}

Ocular abnormalities, such as intraocular tumor (IOT), retinal detachment (RD), posterior scleral staphyloma (PSS), and vitreous hemorrhage (VH), can develop a severe visual disability and even blindness without treatment \cite{stannard2013radiotherapy, ohno2019posterior}. Global assessments in 2020 show that 1.1 billion individuals have ocular diseases, which is expected to rise to 1.8 billion by 2050 \cite{blindness2021vision}. Untimely treatment will increase eye disease and treatment costs, leading to more severe economic losses to patients and families. A recent global financial assessment in 2020 indicates a US\$ 411 billion loss concerning ocular diseases, negatively impacting global medical systems and economic growth \cite{marques2021global}. Hence, early diagnosis and timely treatment are crucial for enhancing visual prognosis and decreasing the financial burden \cite{koh2020novel}. A proper diagnostic instrument is a key to early and quick detection and following cure of ocular anomalies \cite{neupane2018imaging}. As one of the non-invasive measures, ocular ultrasound (OUS) is preferred over biopsy regarding patient comfort and has been widely applied to obtain clinical information for patients with PSS, VH, RD, and vision loss \cite{blaivas2002study, koh2020novel}. Besides, OUS can be utilized in ocular diagnosis to obtain features for detecting an intraocular tumor via the location and lesion extent \cite{neupane2018imaging}.

Nonetheless, ultrasound requires the expertise of ophthalmologists for the accurate and rapid diagnosis of ocular abnormalities. Moreover, ophthalmologists may dispute the diagnosis of ocular abnormalities in clinical practice since explicit or implicit human factors, such as lack of experience and tiredness, may cause the diagnosis to be subjective and prone to error \cite{yoonessi2010bedside}. Therefore, non-biased and automated recognition algorithms are desirable to provide a precise and efficient diagnosis of ocular ultrasound images.

\begin{figure*}[ht]
  \centering
  \includegraphics[width=0.85\textwidth]{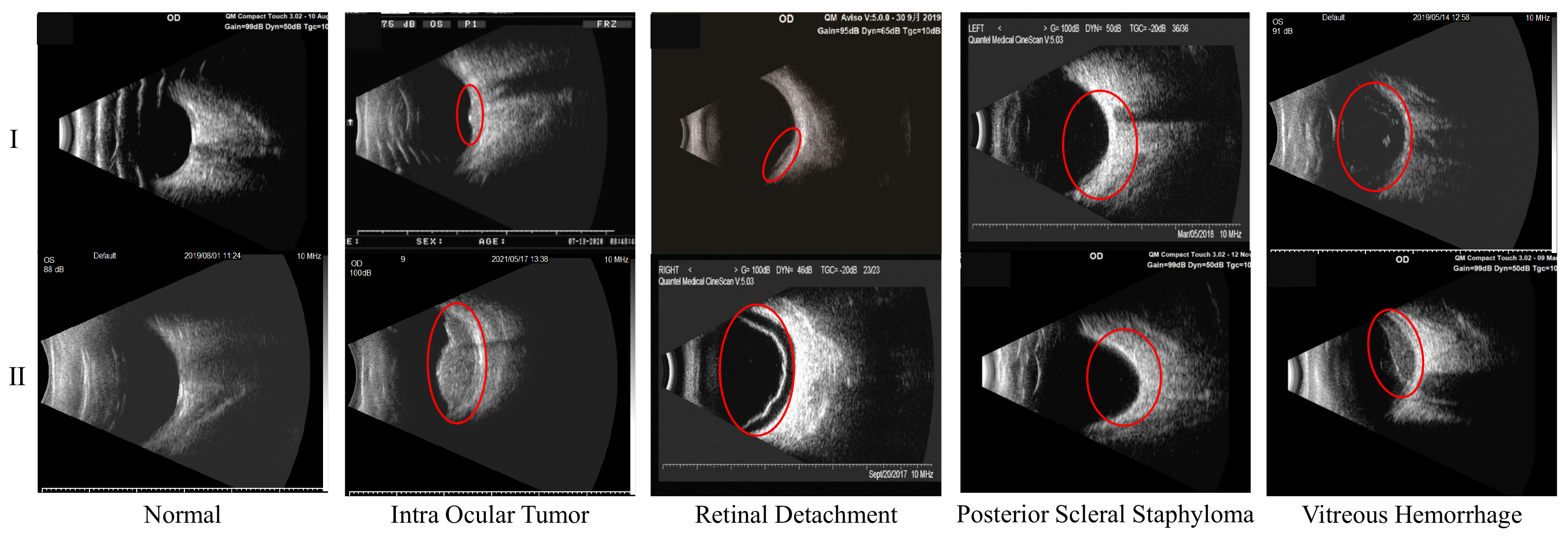}
  \caption{Visualization of our ZJU Ocular Ultrasound Dataset (ZJUOUSD). The abnormal ocular regions are marked via red ellipses. The inconspicuous inter-class variations are illustrated in the first row, where only tiny lesions are present. The significant intra-class variations can be seen comparing the first and the second row.}
  \label{fig:ZJUOUSD_data}
\end{figure*}

However, the diagnosis of ocular abnormalities is a challenging task for existing works due to many intrinsic complexities:
\textbf{1)} As shown in Fig. \ref{fig:ZJUOUSD_data}, the classification performance of models is likely to be severely affected by the irrelevant background, similar spatial features, subtle differences, and contrast diversity between OUS samples, which means it is a challenging fine-grained categorization task.
\textbf{2)} There are often multiple abnormal ocular regions in an image, and they are often concentrated in one area, as depicted in Fig. \ref{fig_cmz_disentangled}. Thus, the model may learn all lesion features when extracting classification features. It may induce the model to focus on the most salient features, resulting in some small pathological changes that tend to be ignored.
\textbf{3)} Extracting only the entangled lesion features will deteriorate models' generalization performance, neglecting tiny and partial lesions and even wrong categorization.

Recently, plenty of network backbones have been proposed for coarse-grained categorization tasks, including VGG \cite{simonyan2014very}, Inception v3 \cite{ioffe2015batch}, ResNet \cite{he2016deep}, DenseNet \cite{huang2017densely}, and Efficientnet \cite{tan2019efficientnet}. However, they are struggling with fine-grained tasks due to the above challenges. Therefore, it is necessary to design a suitable network for ocular tasks. The research on ocular abnormalities is relatively limited, although many studies are devoted to fine-grained categorization tasks \cite{li2021multiscale,li2021agmb}. The existing attention-based and two-stage approaches \cite{xing2020zoom,li2021multiscale} have achieved promising results in fine-grained categorization tasks. These methods are, however, not suitable for the fine-grained categorization of ocular abnormalities, as the lesion feature disentanglement is far under-explored in these frameworks. The feature disentangled-based approaches are more appropriate than those without considering disentanglement for addressing this classification problem, as the former can distinguish the images with only part lesion features. However, classic deep models trained by cross-entropy loss (CE-Loss) to extract coupled features are sub-optimal for the fine-grained OUS recognition task, where tiny lesions are readily neglected. The CE-Loss stimulates models to concentrate on texture correlations instead of semantic and robustly informative features, impeding the feature disentanglement \cite{hong2021disentangling}.

% To handle these challenges, we propose the CDNet based on the hyperspherical contrastive disentangled loss, which is a kind of contrastive learning based on different lesion regions from the same image. Specifically, we utilize the similarity-measuring capacity of the widely-used matrix operation to disentangle parts of lesion features. 

Therefore, we propose disentangling the lesion features in both input and output aspects to handle these challenges. Specifically, to obtain varied views of the lesions for the input disentanglement, the first step of our model is to locate the lesion regions. Previous works \cite{zhao2022local, feng2021two, huang2020rectifying} highly rely on segmentation labels or bounding boxes for further feature disentanglement. Unfortunately, such substantial annotations of lesions are far more expensive and unavailable in our dataset, where merely category labels are accessible. Thus, during the training, CDNet firstly extracts multi-scale features and then locates the lesion regions via the WSLL module, without relying on strong lesions but only on the category labels. Based on these weak lesion locations from the WSLL, we devise a novel CMZ training strategy incorporated into the CDNet to generate views with the tiny lesions for training. Specifically, we employ the CMZ strategy to generate multiply semantic consistency and challenging views, facilitating the decomposition of features.

Nonetheless, results in section \ref{sec:Experiments} demonstrate that decoupling features only in inputs result in sub-optimal performance for fine-grained recognition of the OUS task due to the absence of output disentanglement. Therefore, we propose a novel loss, namely HCD-Loss, based on contrastive learning for the output feature disentanglement. Specifically, we utilize the similarity-measuring capacity of the widely-used matrix operation to disentangle features of lesion parts. To the best of our knowledge, our work is the first to investigate the fine-grained image categorization (FGIC) challenge via both input and output aspects.

More details of our CDNet are depicted in Fig. \ref{fig_pipeline}. The main contributions of our work can be summarized as follows:

\begin{enumerate}
	\item A novel dual-branch CDNet is designed to solve the FGIC task of ocular ultrasound. The CDNet tackles the FGIC task in a novel feature disentanglement paradigm that consists of input and output disentanglement.
	\item We propose a contrastive multi-zoom (CMZ) strategy to promote input feature disentanglement. The CMZ strategy can generate challenging views of lesion regions for the training.
% 	We devise a weakly-supervised lesion localization module to collaborate the disentanglement procedure with only category labels.
	\item We devise a hyperspherical contrastive disentangled loss (HCD-Loss) to boost output feature disentanglement. The HCD-Loss can facilitate the CDNet to extract discriminative and decoupled features for the FGIC task.
	\item We build the ZJUOUSD, a large-scale ocular abnormal images benchmark in ultrasound, which can advance fine-grained OUS abnormalities categorization in bedside practice.
%To our best knowledge, this is the first work that designs a fine-grained and feature disentangled framework for ocular abnormalities in ultrasound.
\end{enumerate}

% \begin{enumerate}
% 	\item A novel dual-branch CDNet is developed to solve the fine-grained image categorization task of ocular ultrasound, utilizing attention maps and the CMZ strategy to zoom in on lesion regions and decompose local regions.
% 	The attention mechanism and CMZ strategy reduce the reliance on strong supervision and induce the input feature disentanglement.
% 	\item A novel HCD-Loss is proposed to boost the feature disentanglement in the output feature. The CDNet utilizing HCD-Loss can extract discriminative and further decoupled features for FGIC.
% 	\item A large-scale benchmark of the ZJUOUSD is built for fine-grained OUS image categorization. The efficacy of the main components is validated on the ZJUOUSD. Meanwhile, extensive experiments on two public chest X-ray (CXR) datasets demonstrate that the proposed CDNet outperforms state-of-the-art CXR classification methods.
% \end{enumerate}

\section{Related Works}
\label{sec:Related_Work}

We propose a contrastive disentangled network to categorize OUS images in this work. Our method mainly involves the four fields: ocular ultrasound recognition, attention mechanism, contrastive representation learning, and feature disentanglement. Thus, the most related works in these fields are reviewed in this section.

\subsection{Ocular ultrasound abnormalities recognition}

\subsubsection{Traditional methods for OUS task}
An accurate and timely diagnosis of ocular abnormities is vital for ocular therapy in the B-scan ultrasound modality. Traditional methods in literature \cite{koh2020novel, gupta2020novel} categorize ocular abnormities via machine learning (ML) and morphology digital image processing. However, the ocular ultrasound diagnosis systems based on traditional methods rely heavily on prior knowledge and doctor-handcrafted features, which hinders the improvement of the ocular diagnosis.

\subsubsection{Deep learning-based methods for OUS task}
Recent deep learning-based methods have improved ocular disease recognition tasks, such as cataract detection and retinal abnormality classification. Wang et al. \cite{wang2021cataract} propose a collaborative monitoring deep learning method to detect cataracts, eliminating the background's influence and enhancing the detection performance. Zheng et al. \cite{zhang2020attention} develop an ensemble attention method (EAM) for cataract recognition, consisting of three classification networks and an object detection network. Like the EAM, Wu et al. \cite{wu2021automatic} propose a cataract detection paradigm based on eyeball detection and a multi-task learning network.

Nevertheless, the above approaches face dilemmas when the location labels of lesions and eyeballs are unavailable since they greatly rely on supervised detection for further categorization. Moreover, these works do not comprehensively explore fine-grained ocular abnormalities, while our work mainly focuses on categorizing tiny ocular lesion problems in a feature-disentanglement paradigm. Many works have investigated weakly-supervised deep models for medical image analysis to alleviate the reliance on strong annotations.

\begin{figure*}[ht]
  \begin{center}
  \includegraphics[width=0.9\textwidth]{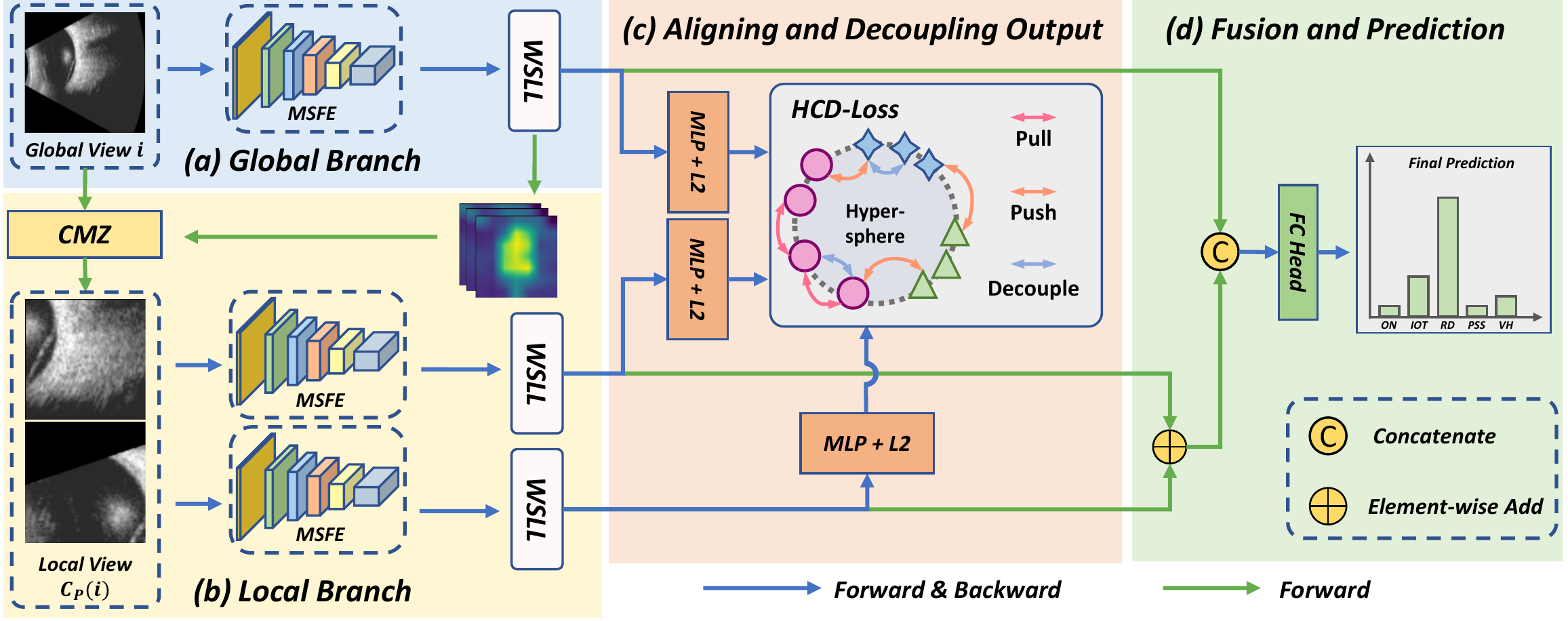}
  \caption{The overall architecture of the proposed CDNet. Global View and Contrastive Local views are fed into a multi-scale feature extractor (MSFE) to capture global and local features for the WSLL. Saliency maps from the global view are utilized to guide the CMZ strategy for the local branch. 
%   The region (a) and (b) represent inputs disentanglement and outputs disentanglement respectively.
  \label{fig_pipeline} 
    }  %note label inside caption
    \end{center}
\end{figure*}

\subsection{Attention Mechanism and Zoom-in Strategy}
Plenty of works have explored the weakly-supervised mechanism for recognition, such as Grad-Cam \cite{selvaraju2017grad}, spatial attention mechanism variants \cite{rao2021studying, li2021multiscale}, and Zoom-in-based architecture \cite{wang2017zoom, dong2018reinforced}. The attention-based Zoom-in \cite{wang2017zoom, dong2018reinforced} models first locate the saliency map and crop and then zoom into it for diagnosis. The key idea to zoom is to generate robust attention to guide the zoom-in process and consequently assist models in capturing more detailed features for subtle lesions. Following similar ideas, Xing et al. \cite{xing2020zoom} constructed an Attention Guided Deformation Network (AGDN) for wireless capsule endoscopy recognition. To generate valuable augmentations and compress background noise, Li et al. \cite{li2021multiscale} propose a multi-scale attention-guided augmentation framework, namely MAG-SD, for the chest X-ray image categorization.

Nonetheless, the above methods neglect that the network is apt to learn coupled features from local regions and significantly degrade, for instance, when merely subtle and partial lesions are visible, as shown in the first row of Fig. \ref{fig:ZJUOUSD_data}. Contrary to the above works, we utilize the attention mechanism to reduce the reliance on strong supervision when decoupling features.

\subsection{Contrastive Representation Learning in Medical Visual Recognition}
Contrastive representation learning has been widely explored recently. It achieves superior performance in self- or semi-supervised training of deep models for medical images \cite{zhang2021ultrasound, zhou2021preservational}. However, contrastive learning without considering relationships among samples impedes the ability of contrastive loss. Combining supervised learning and contrastive learning, Khosla et al. \cite{khosla2020supervised} proposed a SupCon Loss for natural image categorization. Hu et al. \cite{hu2021semi} developed a supervised local contrastive loss to leverage limited pixel-wise annotation for cardiac segmentation following the SupCon \cite{khosla2020supervised}. Chartsias et al. \cite{chartsias2021contrastive} proposed using contrastive learning to mitigate the labeling bottleneck for view classification of echocardiograms. The SupCon demonstrates the potential to facilitate training. However, directly implementing SupCon in medical FGIC tasks fails to work well, as shown in Table \ref{tb:ab_study} due to the intrinsic complexities of the FGIC dataset, which is consistent with the discovery in \cite{cole2021does}. One of the solutions to handle the FGIC task is to expand further the positive and the negative sets for contrastive learning. However, the SupCon highly relies on task-specific and strong augmentations, impeding its capacity in the medical domain, as proven in \cite{khosla2020supervised, chartsias2021contrastive, hu2021semi}. Moreover, none of the above works focus on the feature disentanglement in the ocular FGIC problem.

\subsection{Feature Disentanglement in Medical Visual Recognition}
By dividing independent features into different components, the feature disentanglement mechanism aims at reducing feature coupling with each other for visual tasks\cite{pezeshki2021gradient, hong2021disentangling, xu2021variational, zhao2022local}. Previous works try to capture features in latent space utilizing adversarial learning \cite{pei2021disentangle} or auto-encoder mechanism \cite{cheng2021multimodal}, which are apt to fall into mode collapse and sensitive to hyper-parameters \cite{bau2019seeing}. These works lack the explicit feature decomposition (e.g., lesion regions), leading to a sub-optimal disentangled feature for the downstream task (e.g., FGIC task). Explicitly decomposing inputs, Zhao et al. \cite{zhao2022local} devised a local and global feature disentangled network to identify the benign and malignant thyroid nodules. Feng et al. \cite{feng2021two} employs the triplet loss \cite{schroff2015facenet} and proposed a two-stream framework consisting of a recognition stream and a classification stream for the diagnosis of vertebral compression fracture. To tackle the training instability due to misleading correlation between noises and lesions, Huang et al. \cite{huang2020rectifying} proposed decoupling the correlation using supervision from the bounding box. However, the above works highly rely on the supervision of segmentation labels that are unavailable in our dataset.

In this study, inducing feature disentanglement via the CMZ strategy and the HCD-Loss is essential to advance diagnostic performance. The CMZ strategy caters to offering challenging views for feature disentanglement without expensive annotations, which is distinct from the previous works. The HCD-Loss devotes itself to the output feature disentanglement for better lesion recognition collaborating with the CMZ strategy. Our CDNet simultaneously considers both the input and the output disentanglement that distinguishes itself from previous works while achieving a promising result for the OUS task.

% To the best of our knowledge, we are the first to investigate the large-scale fine-grained image categorization problem of ocular ultrasound diagnosis via deep learning mechanism. 

\section{Methodology}
\label{sec:Methodology}

\subsection{Overview of Our Method }
% As shown in Fig. \ref{fig_pipeline}, the CDNet consists of four parts, including the global branch, local branch, representation output and the fusion and prediction. CDNet first extracts multi-scale features of global views during training and then marks saliency regions (e.g., lesions) via WSLL in a weakly-supervised manner. Subsequently, the CMZ strategy considering the location of saliency regions, generates multi-zoom local views of the global views to reduce the coupling among views. The CDNet takes these disentangled views as the second inputs and outputs lesion-specific features. Three shared-weights multi-layer perceptrons (MLP) armed with L2-normalization serve as the project network for embedding these features in a hypersphere space. The HCD-Loss is designed further for pulling, pushing, and decoupling for varied views on the hypersphere to obtain disentangled and robust features. A fully-connected categorization head (FC Head) is trained on top of the frozen backbone via a CE-Loss. Concatenated features are fed to the FC head to make image-level predictions. Notably, only the central region of the global view marked by the attention map is concatenated with the global feature when testing, which is not depicted in Fig. \ref{fig_pipeline} for simplicity.

As shown in Fig. \ref{fig_pipeline}, our framework consists of four parts, including the (a) global branch, (b) local branch, (c) aligning and decoupling output, and (d) fusion and prediction. The MSFE in (a) first extracts multi-scale features of global views during training and then marks saliency regions (e.g., lesions) via the WSLL in a weakly-supervised manner. Subsequently, the CMZ strategy considering the location of saliency regions in the local branch generates multi-zoom local views of the global views to reduce the coupling among views. The MSFEs in the local branch take these disentangled views as the second inputs and outputs of lesion-specific features. In part (c), three shared-weights multi-layer perceptrons (MLP) armed with L2-normalization serve as the project network for embedding these features in a hypersphere space. The HCD-Loss promotes the global and local branches to pull, push, and decouple varied views to obtain aligned and disentangled features. In part (d), a fully-connected categorization head (FC Head) is trained on top of the frozen backbone (i.e., the MSFE with WSLL) via a CE-Loss. Concatenated features are fed to the FC head to make image-level predictions. Notably, in the test time, the local branch only zooms into the attention center of the global view. It extracts local features for fusion and prediction, which is not depicted in Fig. \ref{fig_pipeline} for simplicity.

We will introduce the main modules of our method in the following paragraphs, including the WSLL module, CMZ strategy, and HCD-Loss.
% Notably, only the central region of the global view is concatenated with the global feature when testing, which is not depicted in Fig. \ref{fig_pipeline} for simplicity.

\subsection{Weakly-Supervised Lesion Localization Module}
\begin{figure}[ht]
  \begin{center}
  \includegraphics[width=0.48\textwidth]{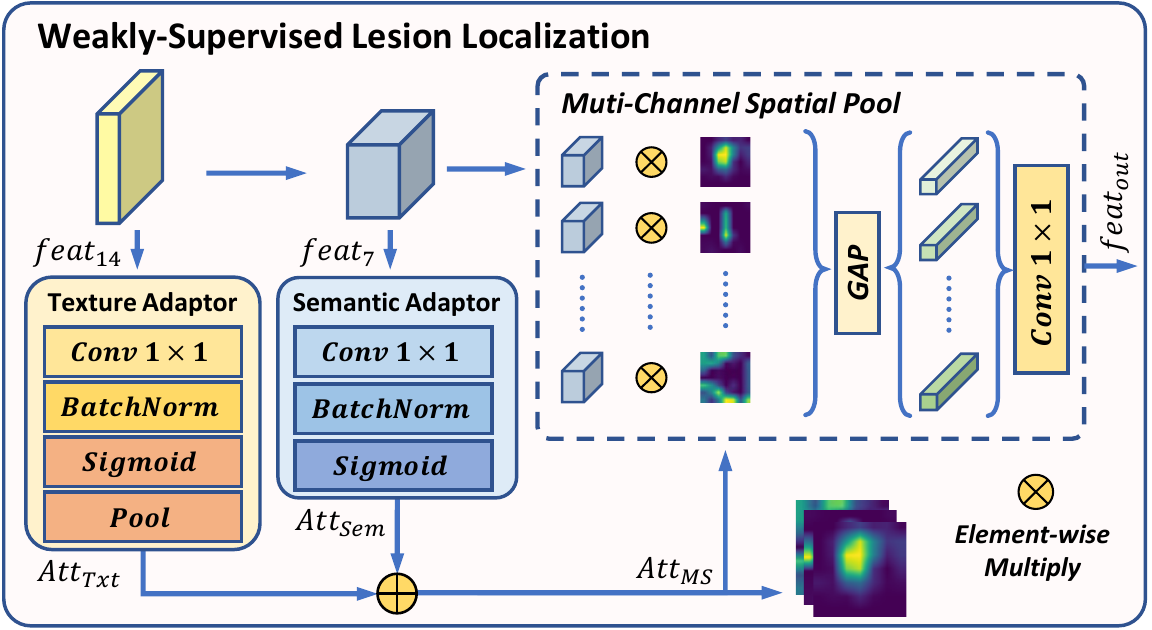}
  \caption{The architecture of the proposed WSLL. 
  % The $M$ is the channel number of attention maps.
  \label{fig_module_WSLL} 
    }  %note label inside caption
    \end{center}
\end{figure}

In image categorization tasks, mere category annotations are available, and lots of works have explored the locating mechanism \cite{selvaraju2017grad, rao2021studying, li2021multiscale, wang2017zoom, dong2018reinforced}. Aiming to utilize feature maps from hierarchical scales, we propose the WSLL module operated on multi-stages of the backbone, including the feature maps $feat_{14}$ and $feat_{7}$ with sizes of $14 \times 14$ and $7 \times 7$. However, a direct stacking of features is not an optimal strategy to capture semantic features, as the literature proves \cite{wang2018deep}. 
%  Nonetheless, most of them confront the problems of low contrast and background noises of the ultrasonic images \cite{wang2022uncertainty, pan2022two}, which hinders the capacity of models to capture tiny differences for FGIC task.
Feature maps from different scales are selected as Texture Adaptor and Semantic Adaptor inputs, respectively, to extract hybrid spatial attention maps. Moreover, we propose a Multi-Channel Spatial Pooling (MCSP) for weighting and squeezing the features, as illustrated in Fig. \ref{fig_module_WSLL}. The MCSP takes the semantic features as input, weights via the multi-stage spatial attention ${Att}_{MS}$ then pool them. The texture spatial attention $Att_{Txt}$, the semantic spatial attention $Att_{Sem}$, and multi-stage spatial attention ${Att}_{MS}$ are calculated by Eq. (\ref{eq:wsll_att_txt}), Eq. (\ref{eq:wsll_att_sem}), and Eq. (\ref{eq:wsll_att_ms}), respectively.
\begin{align}
    \label{eq:wsll_att_txt} &{Att}_{Txt}&  &=& &{ Pool(\sigma(BN(Conv_{1 \times 1}({feat}_{14})))) }& \\
    \label{eq:wsll_att_sem} &{Att}_{Sem}&  &=& &{ \sigma(BN(Conv_{1 \times 1}({feat}_{7}))) }& \\
    \label{eq:wsll_att_ms} &{Att}_{MS}&   &=& &{{Att}_{Txt} + {Att}_{Sem}}&
\end{align}

\noindent where $Pool$, $\sigma$, $BN$ represent average pooling, sigmoid activation, and batch normalization, respectively. Thus, the output feature ${feat}_{out}$ of our WSLL is summarised as Eq. (\ref{eq:wsll_func}).
\begin{equation}
\label{eq:wsll_func}
\begin{aligned}
    {feat}_{out} = {{MCSP}({feat}_{7}, {Att}_{MS})} 
\end{aligned}
\end{equation}

\noindent where ${MCSP}$ represents the MCSP pooling operator. The WSLL can locate lesions in a weakly-supervised manner for further lesion decomposition with the CMZ strategy.
% drl: 这里有一个超参数32, 参考的MAG-SD, 或许需要增加消融实验检验最佳值

\subsection{Contrastive Multi-Zoom strategy}

% Zn, 即缩放次数在这里并未仔细进行消融实验，而是默认选择了2
\begin{algorithm}
	\caption{Contrastive Multi-Zoom strategy} \label{al:CMZ_strategy}
    \KwIn{\\
    \setlength{\parindent}{18pt} Image $\mathcal{I}$, Multi-Stage Spatial Attentions ${Att}_{MS}$, 
    Zoom times ${Z}_{n}$, Threshold of Activation $k$, Parameter of $\beta$ Distribution  $\alpha$ , Crop Scale ${C}_{s}$, Crop Ratio ${C}_{r}$.
    }

	${Att}_{S} \gets Average({Att}_{MS})$                          \tcp{Squeeze spatial attention maps along the channel dimension.}
	$B \gets Gen\-BBox({Att}_{S}, k)$                              \tcp{Generate Bounding Box for the attention via convex hull}
	
	\For{${z}$ $\gets$ $1$ $to$ ${Z}_{n}$}{ \tcp{Zoom ${Z}_{n}$ times via Contrastive Crop}
	        \{ $x_z$, $y_z$, $h_z$, $w_z$ \} $\gets$ $ContrastiveCrop(\mathcal{I}, {Att}_{S}, B, \alpha)$
	}
	$\{ X, Y, H, W \} \gets \{ \{ {x_{z}}\}_{z=1}^{{Z}_{n}} , \{{y_{z}}\}_{z=1}^{{Z}_{n}}, \{{h_{z}}\}_{z=1}^{{Z}_{n}}, \{{w_{z}}\}_{z=1}^{{Z}_{n}}  \}$ \tcp{Set regions to zoom}
	$\{ c_{z} \}_{z=1}^{{Z}_{n}} \gets Zoom \& Interpolate(\mathcal{I}, X, Y, H, W)$       \tcp{Zoom in and resize the views}
	\KwOut{contrastive multi-zoom views $\{ c_{z} \}_{z=1}^{{Z}_{n}} $}
	% 	$\{ {F_{z}}\}_{z=1}^{{Z}_{n}}  \gets Forward(\{{\mathcal{I}_{z}}\}_{z=1}^{{Z}_{n}})$            \tcp{Zoom in forward}
    % 	$\{ {r_{z}}\}_{z=1}^{{Z}_{n}}  \gets L2Norm({\{F_{z}\}_{z=1}^{{Z}_{n}}})$                       \tcp{$L_{2}$ Normalize the representations along the feature dimension}
    % 	\KwOut{contrastive multi-zoom representations $\{ {r_{z}}\}_{z=1}^{{Z}_{n}} $}
\end{algorithm}

\subsubsection{Motivation of the CMZ strategy}
Extracting disentangled features can enhance recognition as not all discriminative regions (e.g., lesions) appear \cite{zhao2022local, han2020learning, zhang2021knowledge}. Intuitively, the various lesion scope induces significant intra-class variance. For instance, the VH images in the first and second row are apt to disturb the training of coarse-grained networks, as depicted in Fig. \ref{fig:ZJUOUSD_data}. Our insight is that the categorization capacity of networks can be enhanced if partial lesion regions are recognized well, as it is a frequent scene in the OUS task. More specifically, low-overlapping and disentangled inputs are more suitable for partial lesion recognition than the entangled ones. Nonetheless, the widely-used attention-guided Center Zoom mechanism only zooms in on the whole suspicious area, which increases the difficulties of feature decomposition and the probability of ignoring subtle regions \cite{zhang2021knowledge, zhao2022local}. We delicately designed our CMZ strategy for reducing the intersection of the local-branch inputs inspired by Contrastive Crop \cite{peng2022crafting}, which is proposed first in self-supervised learning (SSL) for data augmentation.

\subsubsection{Entangled and disentangled lesions}
\begin{figure}[ht]
   \begin{center}
   \includegraphics[width=0.45\textwidth]{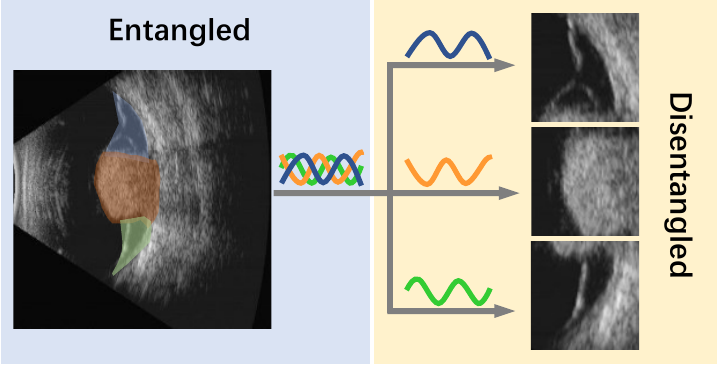}
    \caption{Visualization of the entangled lesions and the disentangled lesions. The regions masked by varied colors represent different parts of lesions.
    \label{fig_cmz_disentangled} 
    }  %note label inside caption
    \end{center}
\end{figure}

As depicted in Fig. \ref{fig_cmz_disentangled}, the entangled views can be disentangled into varied components, miming the circumstance when only partial lesions are visible. Therefore, the MSFE is forced to recognize tiny and partial regions while extracting decoupled features from varied views.

Specifically, the CMZ strategy generates multi-zoom views away from the center of the attention while maintaining a little focus on the most discriminative region. More details are described in the Algorithm. \ref{al:CMZ_strategy}. Our CMZ strategy can significantly reduce the overlapping between cropped views, simulating the condition when only tiny lesions are present in the OUS image. For another, such simulation can facilitate the decomposition of features in the input aspect, thus promoting recognition performance. Center Zoom replaces the Contrastive Crop for the test time since it is more important to diagnose the most salient regions for the final diagnosis. Notably, only the CMZ strategy is not enough for an optimal feature disentanglement of the FGIC task, as proven in section \ref{sec:Experiments}.
% As demonstrated by the attention map and corresponding green box in Fig. \ref{fig_cmz_disentangled}, our CDNet can precisely locate the detached retina in column (a) and (b), i.e., global and the attention center.
% As such, lesion features have a larger impact on the global feature embeddings, thus enabling better inspection for tiny abnormalities.

\subsection{Hyperspherical Contrastive Disentangled Loss}

\begin{figure}[ht]
  \begin{center}
  \includegraphics[width=0.4\textwidth]{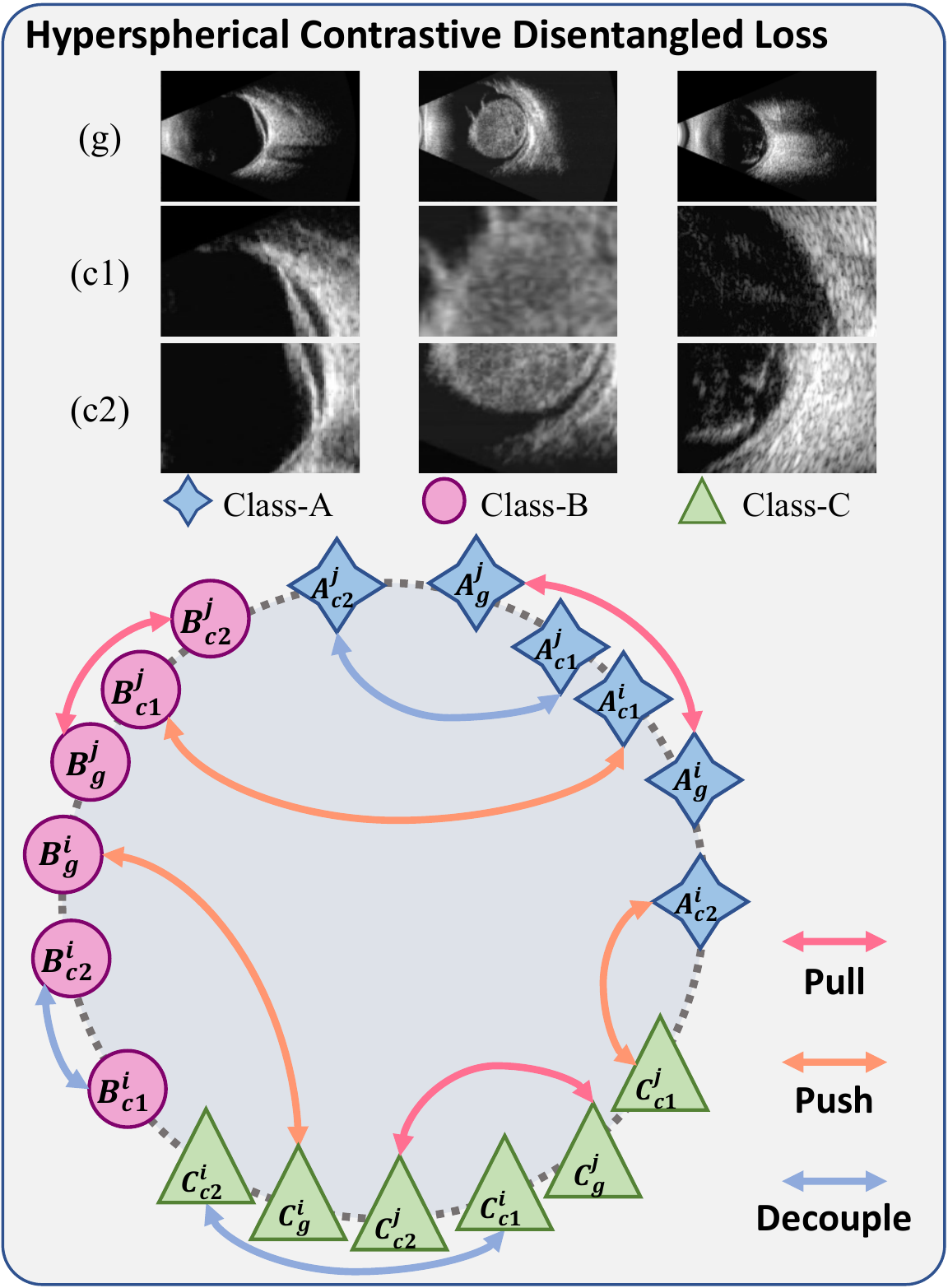}
  \caption{The illustration of the proposed HCD-Loss. Only three categories and partial relationships (indicated by the double arrow) are depicted for simplicity. The row (g) represents the global views, while the row (c1) and (c2) are views that are generated by the CMZ strategy.
  \label{fig_module_HCD_Loss} 
    }  %note label inside caption
    \end{center}
\end{figure}

% Widely used cross-entropy loss (CE-Loss) tends to concentrate on the most salient areas, which is harmful to disentangling the extracted views\cite{zhang2021knowledge, khosla2020supervised, zbontar2021barlow}.
% Unlike the CE-Loss that minimizes the cross-entropy between the models' predictions and the corresponding ground truth labels
\subsubsection{Motivation of the HCD-Loss}
Matrix operation is one of the cornerstones of current deep learning-based methods. It has been widely employed in many visual tasks due to its depicting capacity of relationships between entities \cite{huang2019ccnet, khosla2020supervised, hou2019cross, zbontar2021barlow, chen2021crossvit}. Based on these observations, we propose utilizing the matrix operation's fantastic similarity-measuring capacity for pulling, pushing, and decoupling the different terms. Specifically, to stimulate robust and decoupled feature representation, we devise a novel loss function, HCD-Loss. The HCD-Loss aims to pull together clusters of points from the same category, push apart clusters of points from different categories, and decouple features from the same global views. As depicted in Fig. \ref{fig_module_HCD_Loss}, views (e.g., ${A}_{c2}^{j}$ and ${A}_{c1}^{j}$) from the same categories are pulled together and views (e.g., ${B}_{g}^{i}$ and ${C}_{g}^{i}$) from varied categories are pushed apart, while views (e.g., ${B}_{c1}^{i}$ and ${B}_{c2}^{i}$) from the same image are forced to decouple with each other.

We denote our multi-scale feature extractor as $f(.)$ and the MLP with $L2 Norm$ as $g(.)$. Our goal for the FGIC task is to learn a fine feature embedding network from the training dataset $D=\{(x_i, y_i)\}_{i=1}^{N}$, where $N$ represents the number of our training set samples. The $f$ extracts a high-dimension (e.g., 2048) feature from the input $x_i$. Then $g$ projects the feature to an L2-normalized d-dimensional (e.g., 128) embedding $r_{z}$, such that $r_{z}=g(f(x_i)) \in \mathcal{R}^d $ and falls on the unit hypersphere. Let $\mathcal{I}$ denote the randomly sampled batch images during training from $D$, $\mathcal{P}(i)$ denote the index set of all positive samples of the anchor sample $i$ (i.e., samples with the same category as $i$). The negative pairs in the batch are images with distinct labels to the anchor sample $i$, the indices of which can be denoted as $\mathcal{N}(i)$. Formally, $\mathcal{P}(i)$ and $\mathcal{N}(i)$ are denoted as Eq. (\ref{eq:P_for_supcon_loss}) and Eq. (\ref{eq:N_for_supcon_loss}). One naive approach to utilizing contrastive learning for our task is directly conducting image augmentations to a batch and applying contrastive loss. However, such a procedure orienting self-supervised manners neglects the multi-zoomed views (e.g., row (c1) and (c2) in Fig. \ref{fig_module_HCD_Loss}) from the batch and increases the augmentation processing burden.
\begin{align}
    \label{eq:P_for_supcon_loss} &\mathcal{P}(i)& &=& &\{ j \mid \forall j \in \mathcal{I}, y_j = y_i, and\ j \neq i \}& \\
    \label{eq:N_for_supcon_loss} &\mathcal{N}(i)& &=& &\{ j \mid \forall j \in \mathcal{I}, y_j \neq y_i \}&
\end{align}

\subsubsection{Extending the positive and the negative samples}
To extend samples and facilitate aligning embeddings, we propose to utilize the global views and the corresponding multi-zoomed views. Specifically, the $\mathcal{C}_{P}(i)$, denoted in Eq. (\ref{eq:C_P_for_supcon_loss}), represents the contrastive multi-zoomed views from $i$. The $\mathcal{C}_{N}(i)$, denoted in Eq. (\ref{eq:C_N_for_supcon_loss}), represents contrastive multi-zoomed views from each view of $\mathcal{N}(i)$. Thus the extended positive samples set $\mathcal{P'}(i)$ and negative samples set $\mathcal{N'}(i)$ can be denoted in Eq. (\ref{eq:P_prime_for_supcon_loss}) and Eq. (\ref{eq:N_prime_for_supcon_loss}). This way, the positive and negative samples are substantially extended for robust feature representation.
% &\mathcal{P}(i)& &=& &\{ j \mid \forall j \in \mathcal{I}, y_j = y_i, and\ j \neq i \}& \\
% &\mathcal{N}(i)& &=& &\{ j \mid \forall j \in \mathcal{I}, y_j \neq y_i \}& \\
\begin{align}
    \label{eq:C_P_for_supcon_loss} &\mathcal{C}_{P}(i)& &=& &\{ c \mid c = CMZ(i, {Att}_{s}^{i}, B(i), z) \}_{z=1}^{{Z}_{n}}& \\
    \label{eq:C_N_for_supcon_loss} &\mathcal{C}_{N}(i)& &=& &\{ c \mid c \in \mathcal{C}_{P}(j), \forall j \in \mathcal{N}(i) \}& \\
    \label{eq:P_prime_for_supcon_loss} &\mathcal{P'}(i)& &=& &\mathcal{P}(i) \cup \mathcal{C}_{P}(i)& \\ 
    \label{eq:N_prime_for_supcon_loss} &\mathcal{N'}(i)& &=& &\mathcal{N}(i) \cup \mathcal{C}_{N}(i)&
\end{align}

\subsubsection{Pulling and Pushing for discriminative features}
To extract closely aligned and discriminative representations to all entries from varied classes, we propose the $\mathcal{L}_{pp}(\mathcal{P'}, \mathcal{N'})$. Formally, the $\mathcal{L}_{pp}$ orienting the OUS task is expressed as Eq. (\ref{eq:supcon_loss}).
\begin{equation}
\label{eq:supcon_loss}
\begin{aligned}
    &\mathcal{L}_{pp}( \mathcal{P'} , \mathcal{N'} )& &=& &\sum_{i \in \mathcal{I} }{\frac{-1}{\left| \mathcal{P'}(i) \right|}} \sum_{p \in \mathcal{P'}(i)}{ \mathcal{M}(i, p) }& \\
\end{aligned}
\end{equation}

\noindent where the $\mathcal{M}(i, p)$ represents the similarity between $i$ and $p$. The $\mathcal{M}(i, p)$ is denoted in Eq. (\ref{eq:supcon_loss_M}).
\begin{equation}
\label{eq:supcon_loss_M}
\begin{aligned}
    &\mathcal{M}(i, p)& &=& &log\frac{exp(s_t\cdot cos( \theta_{i}, \theta_{p} ))}{ \sum_{ a \in \mathcal{P'} \cup \mathcal{N'} }{exp(s_t\cdot cos( \theta_{i}, \theta_{a} ))} }& \\
\end{aligned}
\end{equation}

\noindent where the $s_t$ denotes the temperature scaling factor, and the $cos(\theta_{i}, \theta_{j}) = \theta_{i} \cdot \theta_{j}$ denotes cosine similarity, since the $\theta_{i}$ is normalized. Therefore, the negatives summation in the denominator of $\mathcal{M}$ highly contributes to the loss when negatives are continually wrongly categorized, spurring a closer feature embedding.

\subsubsection{Decoupling outputs features}
The $\mathcal{L}_{pp}(\mathcal{P'}, \mathcal{N'})$ results in a more robust clustering of the embedding space than that generated from $\mathcal{L}_{pp}(\mathcal{P}, \mathcal{N})$, as section. \ref{sec:Experiments} proves. Nonetheless, merely $\mathcal{L}_{pp}$ for pulling and pushing will lead to a sub-optimal representation due to neglecting feature disentanglement. Therefore, the cross-correlation matrix between the representation of multi-zoom views is calculated to further decouple different dimensions of features in the aspect of outputs. Formally, the decoupling loss $\mathcal{L}_{dc}$ is defined as Eq. (\ref{eq:cc_matrix_loss}).

\begin{equation}
\label{eq:cc_matrix_loss}
\begin{aligned}
    \mathcal{L}_{dc}(\mathcal{C}_{P}) = d \times \sum_{x}{\sum_{y \neq x}{{cos}^2({V(c_1,x)}, {V(c_2,y)})}} + \\
    \sum_{x}{( cos({V({c_1},x}), {V({c_2},{x}})) - 1 )^2}
    % &\mathcal{L}_{dc}& &=& &\sum_{i}{\sum_{j \neq i}{\mathcal{M}^2}} + d \times \sum_{i}{( \mathcal{M} - 1 )^2}& \\
    % &\mathcal{M}&  &=& &\frac{ \sum_{k}{r_{ki}^{V_1}} \cdot {r_{kj}^{V_2}} }{ \sqrt{ \sum_{k}{ {(r_{ki}^{V_1})}^2 } } \sqrt{ \sum_{k}{ {(r_{kj}^{V_2})}^2 } } }& \\
    % &\mathcal{M}& &=& &\sum_{k}{r_{ki}^{V_1}} \cdot {r_{kj}^{V_2}}&\\
\end{aligned}
\end{equation}

\noindent where the $d \in [0,1] $ represents the balancing hyperparameter, the $c_n \in \mathcal{C}_{P}$ represents the contrastive local views, and the $V(c_n, x) \in \mathcal{R}^{b \times 1}$ represents the dimension vector composed of the $x$th dimension features of a batch of zoomed views. The $V(c_n, x)$ is $L_2$-normalized before calculation, and the $b$ denotes the batch size. Intuitively, the ${cos}^2({V(c_1,x)}, {V(c_2,y)})$ is limited to zero (i.e., orthogonality) for decoupling features in varied dimensions. The $cos({V({c_1},x}), {V({c_2},{x}}))$ is forced to 1, as the specific $x$-dimension features from the same image are expected to be parallel. We only employ the $\mathcal{L}_{dc}$ between contrastive multi-zoom views (i.e., $c_n$) instead of global images since the multi-zoom views are highly entangled with the global while decoupling them tends to overfit background noises. 

\subsubsection{Joint loss function}
Specifically, the outputs of the MLP are used as inputs to the HCD-Loss for training the $f$, as illustrated in Fig. \ref{fig_pipeline}. Therefore, our joint HCD-Loss is denoted as Eq. (\ref{eq:hcd_loss}). 
\begin{equation}
\label{eq:hcd_loss}
\begin{aligned}
    &\mathcal{L}_{HCD}& &=& \mathcal{L}_{pp}(\mathcal{P'}, \mathcal{N'}) + \lambda \mathcal{L}_{dc}(\mathcal{C}_{P})
\end{aligned}
\end{equation}

\noindent where $\lambda$ is the trade-off hyperparameter. We further investigate the selection of $\lambda$ in section \ref{sec:CRC_lambda}. Such that the HCD-Loss can contribute to aligning and decoupling features.

% Furthermore, for hard positives, the effect increases (asymptotically) as the number of negatives does. Eqs. 2 and 3 both preserve this useful property and generalize it to all positives. This implicit property allows the contrastive loss to sidestep the need for explicit hard mining, which is a delicate but critical part of many losses, such as triplet loss.
%What's more, the encoder may be misled by attention areas with noises thus reducing the diversity of generated samples and the semantic consistency between global views and generated ones.
% The semantic-aware localization scheme provides useful guidance to reduce false positive cases, but increases the probability of close appearance pairs due to the smaller operable region.

\section{Experiments and Discussions}
\label{sec:Experiments}

\subsection{Data and Preprocessing}

\subsubsection{Materials}
Although there are some related studies on the categorization task of the OUS images, there is no large-scale, fine-grained labeled public dataset with a standardized collection method available. In this work, ZJUOUSD was built as a large-scale dataset based on the OUS images collected from The Second Affiliated Hospital, Zhejiang University School of Medicine, Hangzhou, China. All images are annotated and examined by experienced ophthalmologists. Our proposed CDNet is validated on our ZJU Ocular Ultrasound Dataset (ZJUOUSD), whose more details are depicted in Table \ref{tb:ZJUOUSD_table}. 

\begin{table}[ht]
\scriptsize
\centering
\begin{center}
\caption{The distribution details of the ZJUOUSD.
 \label{tb:ZJUOUSD_table} }

\begin{tabular}{c|cccccc}
\Xhline{1pt}
Category        & Normal & IOT   & RD   & PSS   & VH    & Total \\ \hline
Train           & 656    & 580   & 848  & 681   & 956   & 3721  \\
Validation      & 90     & 90    & 90   & 90    & 90    & 450   \\
Test            & 186    & 167   & 235  & 193   & 261   & 1042  \\ \hline
Total           & 932    & 837   & 1173 & 964   & 1307  & 5213  \\ \Xhline{1pt}
%Proportion (\%) & 17.88  & 16.06 & 22.5 & 18.49 & 25.07 & 100   \\ \Xhline{1pt}
\end{tabular}

\end{center}
\end{table}

Furthermore, the generalization ability of CDNet is evaluated on two public and widely-used chest X-ray FGIC datasets: Chest X-Ray Images for Classification Dataset (CXRD) \cite{kermany2018identifying} with 5856 cases and COVID-19 RADIOGRAPHY DATABASE (COVID-RD) \cite{rahman2021exploring, chowdhury2020can} with 4035 cases, whose more details are shown in Table \ref{tb:CXRD_and_COVID-RD_table}. The CXRD is taken from patients with normal lungs (normal), bacterial pneumonia (BP), and viral pneumonia (VP). Following the previous work \cite{lv2021cascade} for chest x-ray categorization, the initial test set is sub-divided into a valuation subset and a test subset in this work. The COVID-RD is initially composed of 3616 COVID-19 positive cases (C19), 6012 lung opacity cases, 10192 normal cases (Normal), and 1345 viral pneumonia (VP) CXR images. Following the division protocol of literature \cite{rundo2021advanced}, we select C19, Normal, and VP for analysis, as depicted in Table \ref{tb:CXRD_and_COVID-RD_table}.

\begin{table}[ht]
\caption{The distribution details of the CXRD and COVID-RD. The BP class only exists in the CXRD, while the C19 only exists in the COVID-RD.
 \label{tb:CXRD_and_COVID-RD_table} }
\scriptsize
\centering
\begin{tabular}{c|ccccc}
\Xhline{1pt}
Dataset                   & Class           & Normal & BP (C19)  & VP    & Total \\ \hline
\multirow{4}{*}{CXRD}     & Train           & 1341   & 2530      & 1345  & 5216  \\
                          & Validation      & 125    & 121       & 74    & 320   \\
                          & Test            & 125    & 121       & 74    & 320   \\ \cline{2-6} 
                          & Total           & 1591   & 2772      & 1493  & 5856  \\ \hline
                          %& Proportion (\%) & 27     & 47       & 26    & 100   \\ \hline
\multirow{4}{*}{COVID-RD} 
                          & Train           & 940    & 940       & 940   & 2820  \\
                          & Validation      & 200    & 200       & 200   & 600   \\
                          & Test            & 205    & 205       & 205   & 615   \\ \cline{2-6} 
                          & Total           & 1345   & 1345      & 1345  & 4035  \\ \Xhline{1pt} 
                          %& Proportion (\%) & 33.33  & 33.33    & 33.33 & 100   \\ \Xhline{1pt}
\end{tabular}

\end{table}

\subsubsection{Preprocessing and Augmentation}
The original ultrasound images have small but misleading icons, including details about the patients and the imaging devices' parameters, confusing the model's optimization. Thus we remove background noises, crop the central region from original ultrasound images via setting a specific cropping location for each imaging device and resize all of them to $224 \times 224$ for reasoning, same for CXRD and COVID-RD. Besides, data augmentation is well-known for remitting the impact of overfitting, thus improving the robustness of deep networks since the labeled data is precious in medical FGIC tasks. Specifically, we implement the common data augmentation mechanism via a widely-used tool\cite{info11020125}. Augmentation transformations for each training phase are randomly selected from a pre-defined sequence of transformation lists: Rotation, Flip, Affine transformation, and CLAHE.
  
\subsection{Performance Comparison}
\subsubsection{Network Training and Evaluation metrics}
Our experiments were based on the PyTorch framework and NVIDIA RTX 2080Ti GPUs. We trained our network from scratch for 128 epochs, and the Adam algorithm (momentum = 0.97, weight decay = $5 \times 10^{-4}$) updated the parameters. We adopted a batch size of 64 and set the learning rate as $1 \times 10^{-3}$. To evaluate the classification performance, we adopted the commonly used metrics in this study, including F1-score, Accuracy, Precision, Sensitivity, and Specificity.

% which are computed as Eq. (\ref{eq:metrics_F1}, \ref{eq:metrics_Acc}, \ref{eq:metrics_Pre}, \ref{eq:metrics_Sen}, \ref{eq:metrics_Spe}).
% \begin{equation}
% \label{eq:metrics}
% \begin{aligned}
%     &F1\mbox{-}score&      &=& &\frac{2 \times TP}{2 \times TP + FP + FN}& \\
%     &Accuracy&             &=& &\frac{TP + TN}{TP + TN + FP + FN}& \\
%     &Precision&            &=& &\frac{TP}{TP + FP}& \\
%     &Sensitivity&          &=& &\frac{TP}{TP + FN}& \\
%     &Specificity&          &=& &\frac{TN}{TN + FP}& \\
% \end{aligned}
% \end{equation}

\begin{itemize}
    \item[$\bullet$] F1-score (F1):
    \begin{equation}
    \label{eq:metrics_F1}
        \begin{aligned}
            F1\mbox{-}score  = \frac{2 \times TP}{2 \times TP + FP + FN}
        \end{aligned}
    \end{equation}

    \item[$\bullet$] Accuracy (Acc):
    \begin{equation}
    \label{eq:metrics_Acc}
        \begin{aligned}
            Accuracy    = \frac{TP + TN}{TP + TN + FP + FN}
        \end{aligned}
    \end{equation}

    \item[$\bullet$] Precision (Pre):
    \begin{equation}
    \label{eq:metrics_Pre}
        \begin{aligned}
            Precision   = \frac{TP}{TP + FP}
        \end{aligned}
    \end{equation}

    \item[$\bullet$] Sensitivity (Sen):
    \begin{equation}
    \label{eq:metrics_Sen}
        \begin{aligned}
            Sensitivity  = \frac{TP}{TP + FN}
        \end{aligned}
    \end{equation}

    \item[$\bullet$] Specificity (Spe):
    \begin{equation}
    \label{eq:metrics_Spe}
        \begin{aligned}
            Specificity  = \frac{TN}{TN + FP}
        \end{aligned}
    \end{equation}
\end{itemize}

\noindent where $TP$, $TN$, $FP$, and $FN$ denote the numbers of true positives, true negatives, false positives, and false negatives, respectively. The area under the receiver operating characteristic curve (AUC) is computed to compare model functionality. The better results are depicted in bold in all tables. Furthermore, a confusion matrix is introduced to depict the distribution of our prediction results of multiple classes.

\subsubsection{Backbone Selection}
In order to show the superiority of the proposed method, we compare the classical coarse-grained categorization networks including VGG\cite{simonyan2014very}, Inception v3\cite{ioffe2015batch}, ResNet\cite{he2016deep}, DenseNet\cite{huang2017densely}, and Efficientnet\cite{tan2019efficientnet} since there is little previous work focusing on the large-scale fine-grained OUS image categorization. As shown in Table \ref{tb:cmp_metrics_ZJUOUSD}, a large-scale benchmark of our ZJUOUSD is reported on classical deep learning-based coarse-category classification networks. We select the Efficientnet-b3 as our backbone, as it achieves the state-of-the-art performance among the classical coarse-grained categorization networks. Our model specializes in the fine-grained OUS visual categorization task compared with classic coarse-grained models.

\subsubsection{Comparison With Other Methods}
\label{sec_Comparison_With_Other_Methods}
To further investigate the fine-grained OUS task, a state-of-the-art fine-grained categorization network MAG-SD \cite{li2021multiscale} based initially on ResNet-50 was re-implemented on our ZJUOUSD. Meanwhile, MAG-SD based on Efficientnet is also reported for a fair comparison. The original MAG-SD achieves a 91.8±1.44\% F1 score on our ZJUOUSD and a 92.52±0.46\% F1 score based on Efficientnet-b3. 

Furthermore, our HCD-Loss is replaced by SupCon Loss \cite{khosla2020supervised} (i.e., CDNet (SupCon) in Fig. \ref{tb:cmp_metrics_ZJUOUSD} ) for investigating the functionality of the feature disentanglement. Significant performance advancement is observed in our CDNet armed with HCD-Loss, which illustrates that only input decoupling is suboptimal for the OUS task. Specifically, our CDNet obtains a 94.13±0.03\% F1 score and 94.4±0.05\% accuracy, which is the highest in the benchmarks of the ZJUOUSD. As shown in Table \ref{tb:cmp_metrics_ZJUOUSD}, our CDNet shows superiority in statistical analysis. Since none of these methods (e.g., MAG-SD) considers feature disentanglement, it is expected. In contrast, our method comprehensively decouples the feature at both the input and the output aspects. % Additionally, our CMZ-Net based on Efficientnet-b0 is reported to investigate the impact of different-scale architectures, where Efficient-b3 is larger than Efficient-b0 concerning network capacity. It can be seen that replacing Efficientnet-b0 with -b3 can significantly improve the categorization performance.

\begin{table*}[ht]
\scriptsize
\begin{center}
\caption{Benchmarks of our ZJUOUSD, where Effi-b3 means Efficientnet-b3 for simplicity.
 \label{tb:cmp_metrics_ZJUOUSD} }
\setlength{\tabcolsep}{3mm}
\begin{tabular}{ccccccc}
\Xhline{1pt}
Model              & F1 (\%)             & Acc (\%)            & Pre (\%)            & Sen (\%)            & Spe (\%)            & AUC (\%)            \\ \hline
VGG16              & 84.54±2.65          & 85.22±2.93          & 85.22±2.62          & 84.31±2.64          & 96.28±0.75          & 96.79±0.50          \\
VGG19              & 81.24±0.23          & 81.96±0.66          & 81.85±0.43          & 81.36±0.03          & 95.51±0.12          & 95.98±0.08          \\
ResNet34           & 89.79±0.13          & 90.34±0.06          & 90.74±0.01          & 89.38±0.15          & 97.56±0.02          & 98.70±0.05          \\
ResNet50           & 88.88±0.79          & 89.32±0.84          & 89.83±0.46          & 88.48±0.81          & 97.30±0.21          & 98.60±0.10          \\
ResNet101          & 85.34±0.18          & 86.24±0.24          & 86.16±0.21          & 85.05±0.17          & 96.56±0.06          & 97.28±0.16          \\
Inception v3       & 90.16±0.20          & 90.50±0.17          & 90.63±0.28          & 89.88±0.21          & 97.61±0.04          & 98.56±0.11          \\
DenseNet121        & 90.43±0.52          & 90.95±0.44          & 91.53±0.22          & 90.01±0.61          & 97.72±0.12          & 98.94±0.09          \\
DenseNet201        & 90.74±0.97          & 91.14±0.94          & 91.49±0.83          & 90.33±1.00          & 97.76±0.24          & 98.95±0.06          \\
EfficientNet-b0    & 90.16±1.01          & 90.69±1.00          & 90.96±0.87          & 89.81±1.06          & 97.67±0.25          & 98.63±0.20          \\
EfficientNet-b1    & 91.05±0.13          & 91.62±0.11          & 91.67±0.14          & 90.73±0.10          & 97.90±0.02          & 98.82±0.03          \\
EfficientNet-b2    & 90.97±0.56          & 91.39±0.55          & 91.60±0.48          & 90.62±0.55          & 97.84±0.14          & 98.80±0.01          \\
EfficientNet-b3    & 91.33±0.35          & 91.81±0.28          & 91.97±0.37          & 91.00±0.35          & 97.95±0.06          & 98.87±0.15          \\ \hline
MAG-SD (original)  & 91.80±1.44          & 92.19±1.36          & 92.34±1.17          & 91.51±1.57          & 98.04±0.35          & 99.09±0.18          \\
% MAG-SD ( Effi-b0)  & 91.38±0.86          & 91.75±0.84          & 91.89±0.71          & 91.15±0.86          & 97.93±0.21          & 98.95±0.16        \\
MAG-SD (Effi-b3)   & 92.52±0.46          & 92.83±0.36          & 92.87±0.53          & 92.31±0.42          & 98.20±0.08          & 99.01±0.06          \\ \hline
% CMZ-Net ( Effi-b0) & 93.52±0.60          & 93.86±0.60          & 93.80±0.48          & 93.37±0.66         & 98.47±0.16        & \textbf{99.31±0.14} \\
CDNet (SupCon)    & 92.37±1.08           & 92.71±1.01          & 93.06±0.61          & 92.12±1.12          & 98.17±0.25          & \textbf{99.35±0.17}  \\
CDNet             & \textbf{94.13±0.03}  & \textbf{94.40±0.05} & \textbf{94.43±0.04} & \textbf{93.93±0.06} & \textbf{98.59±0.02} & 99.18±0.06           \\ \Xhline{1pt}
\end{tabular}

\end{center}
\end{table*}

% drl: 这里是否应该再加一个 backbone 为 ResNet50 的？

Considering the radiological characteristics of the datasets above, we compared our method with the current deep learning-based image-level classification methods \cite{lv2021cascade, rundo2021advanced}, which reveal the preliminary benchmarking results and corresponding state-of-the-art performance. With the assistance of the feature disentanglement, our CDNet surpasses the SEME-DenseNet169, the MAG-SD, and the 3D Non-Local DenseNet on all categorization metrics, as depicted in Table \ref{tb:cmp_metrics_CXRD} and Table \ref{tb:cmp_metrics_COVID_RD}.

\begin{table*}[ht]
\scriptsize
\begin{center}
\caption{Benchmarks of the CXRD, where Effi-b3 means EfficientNet-b3 for simplicity.
 \label{tb:cmp_metrics_CXRD} }

\begin{tabular}{ccccccc}
\Xhline{1pt}
Model                      & F1 (\%)             & Acc (\%)            & Pre (\%)            & Sen (\%)            & Spe (\%)           & AUC (\%)            \\ \hline
Base-VGG19                 & 69.00               & 69.69               & 74.00               & 72.00               & 86.00              & 77.60               \\
Base-ResNet50              & 72.00               & 72.81               & 74.00               & 73.00               & 87.00              & 78.10               \\
Base-DenseNet169           & 77.00               & 77.50               & 80.00               & 80.00               & 90.00              & 78.50               \\
VGG19 (GAP)                & 77.00               & 77.81               & 78.00               & 79.00               & 90.00              & 82.40               \\
ResNet50 (GAP)             & 73.00               & 74.06               & 78.00               & 77.00               & 88.00              & 79.60               \\
DenseNet169 (GAP)          & 80.00               & 80.94               & 82.00               & 82.00               & 91.00              & 85.10               \\
SE-ResNet50                & 81.00               & 81.56               & 81.00               & 93.00               & 91.00              & 85.00               \\
SE-DenseNet169             & 81.00               & 81.87               & 83.00               & 83.00               & 92.00              & 84.20               \\
SEME-ResNet50              & 88.00               & 89.06               & 89.00               & 89.00               & 95.00              & 92.10               \\
SEME-DenseNet169           & 81.00               & 82.81               & 83.00               & 82.00               & 92.00              & 89.00               \\
MAG-SD (original) & 86.70±0.95          & 87.50±1.13          & 86.98±0.70          & 87.90±0.66          & 94.20±0.31         & 96.81±0.37          \\ \hline
CDNet ( Effi-b3)           & \textbf{90.21±0.65} & \textbf{90.93±0.54} & \textbf{90.06±0.43} & \textbf{90.75±1.24} & \textbf{95.61±0.4} & \textbf{97.05±0.05} \\ \Xhline{1pt}
\end{tabular}

\end{center}
\end{table*}

\begin{table*}[ht]
\scriptsize
\begin{center}
\caption{Benchmarks of the COVID-RD, where Effi-b3 means EfficientNet-b3 for simplicity.
 \label{tb:cmp_metrics_COVID_RD} }

\begin{tabular}{ccccccc}
\Xhline{1pt}
Model                      & F1 (\%)             & Acc (\%)            & Pre (\%)            & Sen (\%)            & Spe (\%)            & AUC (\%)            \\ \hline
DenseNet-201               & 96.55               & 97.72               & 97.51               & 95.61               & 98.78               & None                \\
3D Non-Local DenseNet      & 97.06               & 98.05               & 97.54               & 96.59               & 98.78               & None                \\
MAG-SD (original) & 96.97±0.37          & 96.96±0.38          & 97.01±0.36          & 96.96±0.38          & 98.48±0.19          & \textbf{99.76±0.11} \\ \hline
CDNet ( Effi-b3)           & \textbf{98.38±0.33} & \textbf{98.37±0.33} & \textbf{98.39±0.32} & \textbf{98.37±0.33} & \textbf{99.19±0.17} & 99.75±0.17   \\\Xhline{1pt}
\end{tabular}

\end{center}
\end{table*}

\subsection{Ablation Study of Framework Design}
% 先有WSLL，才可以实现病灶定位，才可以进行双阶段网络，才可以改进到CMZ策略，才能用上HCD-Loss

Our method components could be summarized as the WSLL module, CMZ strategy, and HCD-Loss. Each component was studied by evaluating its improvement in categorization performance, which is quantified by the metrics above. The following procedure obtained the performance gain: the proposed model was first trained on a specific dataset with metrics. Then, the single component was substituted or removed and validated the same dataset for a fair comparison. All the illustrated models were trained and evaluated on the same protocol. Components validations were reported in Tables \ref{tb:ab_study}.
% For all the tested models, mean value and standard deviation of F1, Acc, Pre, Sen, Spe and AUC were recorded.

\begin{table*}[ht]
\scriptsize
\begin{center}
\caption{Ablation study of each key component.
 \label{tb:ab_study} }

\begin{tabular}{c|ccc|cccccc}
\Xhline{1pt}
Dataset                   &  WSLL &  CMZ &   HCD-Loss     & F1 (\%)             & Acc (\%)            & Pre (\%)            & Sen (\%)            & Spe (\%)            & AUC (\%)            \\ \hline
\multirow{4}{*}{ZJUOUSD}  &      &   &                    & 91.33±0.35          & 91.81±0.28          & 91.97±0.37          & 91.00±0.35          & 97.95±0.06          & 98.87±0.15          \\
                          &  \checkmark &  &       & 91.74±0.28          & 92.16±0.31          & 92.08±0.53          & 91.60±0.29          & 98.04±0.07          & 99.11±0.14          \\
                          
                          &    \checkmark&\checkmark&     & 93.05±0.45          & 93.38±0.44          & 93.41±0.44          & 92.84±0.38          & 98.34±0.10          & \textbf{99.26±0.04}          \\
                          
                         &    \checkmark&\checkmark&\checkmark   & \textbf{94.13±0.03} & \textbf{94.40±0.05} & \textbf{94.43±0.04} & \textbf{93.93±0.06} & \textbf{98.59±0.02} & 99.18±0.06          \\ \hline
\multirow{4}{*}{CXRD}     & & &     & 72.98±1.67          & 74.17±1.60          & 78.07±1.46          & 75.45±1.64          & 88.09±0.89          & 92.16±1.36          \\
                          
                          &    \checkmark&&              & 84.87±1.21          & 85.52±1.26          & 85.97±0.83          & 86.78±1.05          & 93.47±0.54          & 96.34±0.24          \\
                          
                          &    \checkmark&\checkmark&          & 86.47±1.37          & 87.40±1.26          & 86.58±0.93          & 87.39±1.07          & 94.02±0.51          & 96.66±0.23          \\
                          
                          &    \checkmark&\checkmark&\checkmark     & \textbf{90.21±0.65} & \textbf{90.93±0.54} & \textbf{90.06±0.43} & \textbf{90.75±1.24} & \textbf{95.61±0.40} & \textbf{97.05±0.05} \\ \hline
\multirow{4}{*}{COVID-RD} & & &    & 95.98±0.75          & 95.99±0.75          & 96.02±0.71          & 95.99±0.75          & 97.99±0.38          & 99.61±0.15          \\
                          
                          &    \checkmark& &               & 97.03±0.10          & 97.02±0.09          & 97.14±0.13          & 97.02±0.09          & 98.51±0.05          & \textbf{99.78±0.16}          \\
                          
                          &    \checkmark&\checkmark&        & 97.68±0.09          & 97.67±0.09          & 97.72±0.03          & 97.67±0.09          & 98.83±0.05          & 99.78±0.18          \\
                          
                          &    \checkmark&\checkmark&\checkmark    & \textbf{98.38±0.33} & \textbf{98.37±0.33} & \textbf{98.39±0.32} & \textbf{98.37±0.33} & \textbf{99.19±0.17} & 99.75±0.17          \\ \Xhline{1pt}
\end{tabular}

\end{center}
\end{table*}

\subsubsection{Efficacy of weakly-supervised lesion localization (WSLL)}
Typically, state-of-the-art coarse-grained CNN models suffer from similar and entangled features when dealing with FGIC tasks. Existing works toward the feature disentanglement highly rely on the lesion location, which could be effectively located by our WSLL module. Therefore, we first investigate the functionality of our WSLL, which provides our CDNet with lesions localization under a weakly-supervised protocol for further feature disentanglement. With the assistance of WSLL, the EfficientNet-b3 obtains superior performance on all categorization metrics than the baseline on the three datasets, as depicted in Table \ref{tb:ab_study}. As for the Efficientnet-b3 equipped with WSLL, F1 reached 91.74\% in the ZJUOUSD dataset. Moreover, our CDNet armed with WSLL is capable of capturing global inspection and sensitive to tiny ocular lesions, as demonstrated in Fig. \ref{fig:att_vis_ZJUOUSD}. The abnormal regions are subtle, while our CDNet is capable of detecting the lesions, for instance, the IOT row, where a mere tiny tumor is present.

Notably, the WSLL module is plug-and-play and independent of network architecture and input modality. It is capable of locating the lesions under various circumstances, e.g., x-rays. With the guidance of the WSLL, the proposed CDNet is capable of categorizing OUS images while locating lesion regions based on the saliency maps. Moreover, the quantitive and qualitative results demonstrate the efficacy of our WSLL while depicting the diagnosis consistency with experienced ophthalmologists.

\subsubsection{Efficacy of contrastive multi-zoom (CMZ)}
The attention mechanism emphasized local features that can guide examinations further and provide more interpretable decisions. The vanilla Zoom-in\cite{wang2017zoom, li2021multiscale} model that zooms into the center is employed as the baseline of our CDNet, as depicted in Table \ref{tab:zoom_in_vs_cmz}, where significant performance decline emerge. Such decline is because only zooming into the center field of the saliency map when training hinders the decomposition of extracting features of lesions. The vanilla faces misleading background noises due to weakly-supervised settings.

Contrary to the vanilla, the CMZ strategy generates low overlapping regions for disentangling the lesion regions, mimicking the dilemma that merely partial lesions appear. As illustrated in Table \ref{tb:ab_study}, the CMZ strategy facilitates the classification utility of the two-stage model in all three datasets, demonstrating its potentiality in input feature disentanglement for medical recognition. Moreover, the CMZ strategy serves as an attention-guided augmentation, which generates views consistent with global views in the semantic aspect and minimizes the probability of noise caused by random augmentation.

Decomposing the zoom-in views inputs via the CMZ strategy is able to tackle the bedside condition in which negligible and partial lesions appear, which simulates the circumstance when only partial and tiny lesions are present, as depicted in Fig. \ref{fig_cmz_disentangled}. Attention maps in Fig. \ref{fig:att_vis_ZJUOUSD} illustrate the locating ability of our CDNet for subtle abnormalities. These promising results further illustrate the potentiality of feature decoupling and data augmentation of our CMZ strategy.

\begin{table}[]
\scriptsize
\begin{center}
\caption{The comparison results between the vanilla Zoom-in and our CMZ strategy on ZJUOUSD. Notably, our HCD-Loss is not utilized here. 
\label{tab:zoom_in_vs_cmz} }
\begin{tabular}{c|cccccc}
\Xhline{1pt}
Strategy & F1 (\%)        & Acc (\%)       & Pre (\%)       & Sen (\%)       & Spe (\%)       & AUC (\%)       \\ \hline
Zoom in  & 92.31          & 92.71          & 93.01          & 92.04          & 98.17          & 99.17          \\
CMZ      & \textbf{93.05} & \textbf{93.38} & \textbf{93.41} & \textbf{92.84} & \textbf{98.34} & \textbf{99.26} \\ \Xhline{1pt}
\end{tabular}
\end{center}
\end{table}

\begin{figure}[ht]
  \centering
  \includegraphics[width=0.45\textwidth]{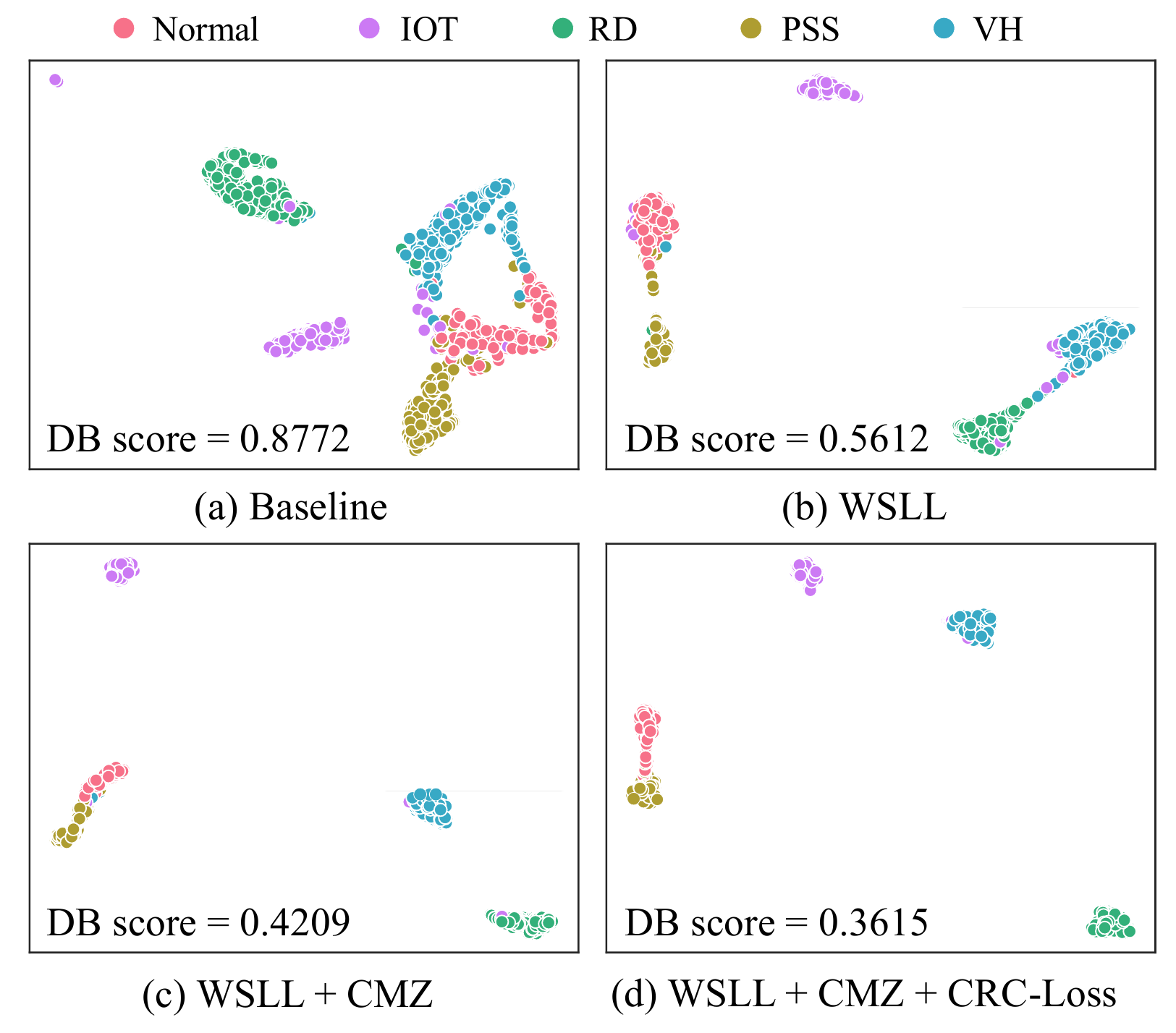}
  \caption{Visualization of the feature distribution of ZJUOUSD via UMAP. The three main variants of our models and the baseline are illustrated. Each variant's davies-Bouldin score (DB score) is also depicted. The minimum DB score, in theory, is zero, and lower values indicate better clustering performance.
  \label{fig:umap_ZJUOUSD} 
    }  %note label inside caption
\end{figure}

\subsubsection{Efficacy of hyperspherical contrastive disentangled loss (HCD-Loss)}
\label{sec:CRC_lambda}
We proposed the HCD-Loss for the output disentanglement to obtain robust and disentangled representations for recognition. Three relationships among views, including pulling, pushing, and decoupling, are explored in our study. The term armed with HCD-Loss achieves promising performance in all three datasets, as depicted in Table \ref{tb:ab_study}. Our experiments indicate that only inputs feature decomposition is apt to be disturbed by classical CE-Loss, which is thus substituted by our novel HCD-Loss. Furthermore, we replace HCD-Loss with SupCon loss\cite{khosla2020supervised} to investigate the disentangled efficacy of the proposed HCD-Loss. Replacing HCD-Loss with the original SupCon loss deteriorates the main metrics, indicating that a straight substitution of SupCon loss without considering local views relationship and feature disentanglement struggles in FGIC tasks, as shown in Table \ref{tb:cmp_metrics_ZJUOUSD}. Specifically, the CDNet with SupCon obtains an average 92.37\% F1 score, far inferior to that of the HCD-Loss. In contrast, our HCD-Loss complements our CMZ strategy regarding the outputs aspect, further demonstrating the importance of the simultaneous feature disentanglement in both the inputs and outputs aspects.

\begin{figure}[ht]
  \centering
  \includegraphics[width=0.45\textwidth]{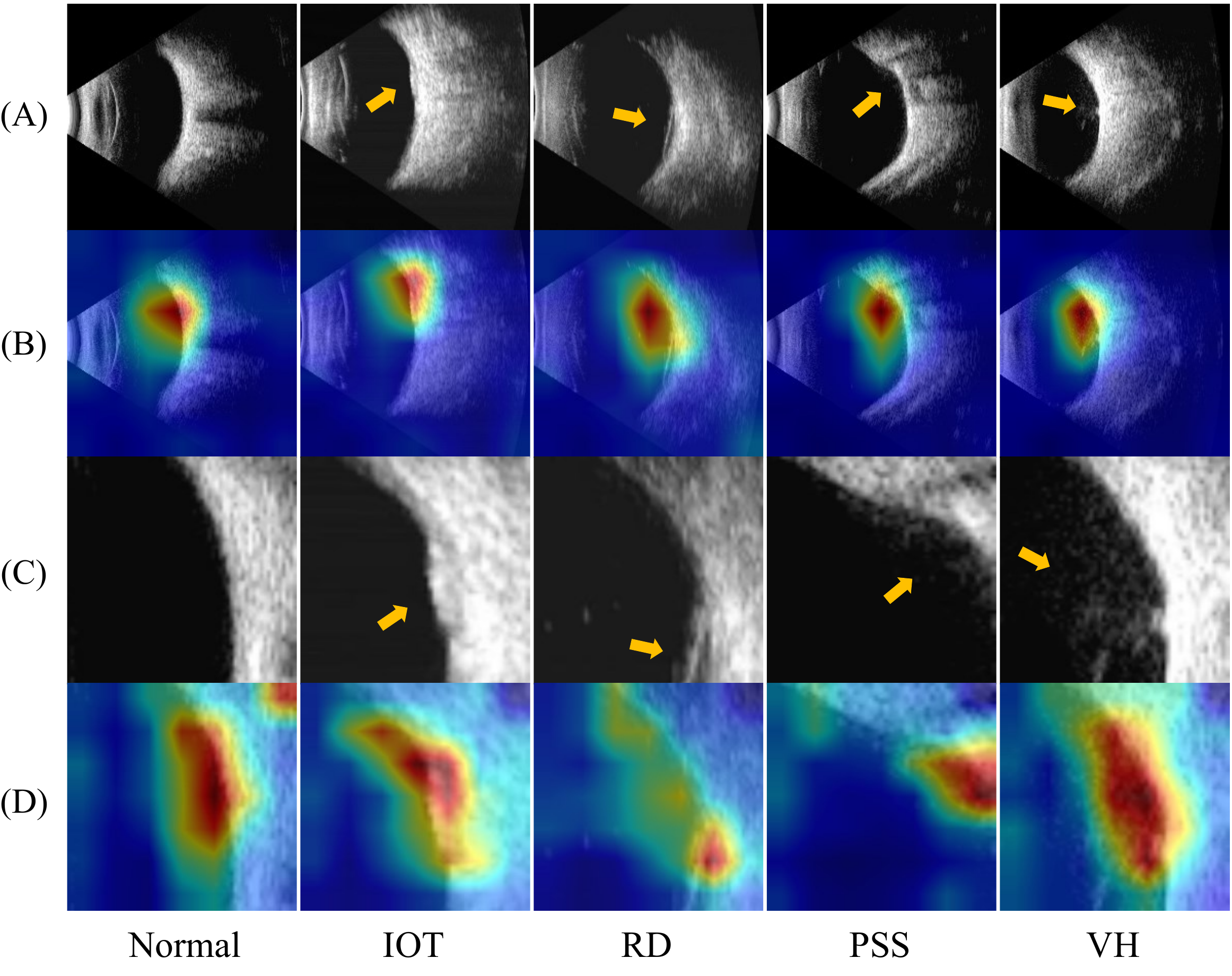}
  \caption{Visualization of the attention heatmaps of ZJUOUSD. The ocular abnormalities are marked with yellow arrows in rows (A) and (C) by experienced ophthalmologists. The brighter the heatmap is, the more critical it illustrates. The (A) and (B) represent the global views and corresponding attention maps. The (C) and (D) represent the local views (at the attention center) and corresponding attention maps.
  }
  \label{fig:att_vis_ZJUOUSD}
\end{figure}

Besides, the global and local branches share the same network and train with different scales of the augmented images, from which the network views different lesions and makes various errors. HCD-Loss serves as an implicit constraint among various views due to the difficulties of self-correcting, promoting the network to extract robust and disentangled representations. Though the global attention did not locate the detached retina well, the local attention precisely located the partial retina for further diagnosis, as illustrated in the RD column in Fig. \ref{fig:att_vis_ZJUOUSD}. Utilizing the common matrix operation promotes the feature decomposition via forcing different-dimension features of different views to be orthogonal and the same-dimension features of different views to be parallel. The HCD-Loss collaborates with the CMZ strategy for feature disentanglement in both the output and the input aspects.

% \begin{figure*}[ht]
%   \centering
%   \includegraphics[width=0.7\textwidth]{imgs/heatmap_vis_CXRD_COVID_RD.pdf}
%   \caption{Visualization of the attention heatmaps of CXRD and COVID-RD.
%   }
%   \label{fig:att_vis_CXRD_COVID_RD}
% \end{figure*}

% \begin{figure*}[ht]
%   \centering
%   \includegraphics[width=0.8\textwidth]{imgs/metrics_line_plots.pdf}
%   \caption{The hyperparameter searching on three datasets for $\lambda$, which is set to 0.02, 0.05, 0.1, 0.2, and 0.4. The $\lambda$ varies on different datasets, where 0.1, 0.2, and 0.05 achieve best performance on three datasets, respectively.
%   }
%   \label{fig:lambda_line_plot}
% \end{figure*}

\subsubsection{Impact of the hyperparameter $\lambda$}
We conduct a series of hyperparameter tuning for better feature disentanglement and representation capacity, validating the trade-off functionality of the $\lambda$ of our HCD-Loss. All settings are frozen except the $\lambda$ of the HCD-Loss for condition control. All metrics above are depicted in Fig. \ref{tab:lambda_ablation}, from which we can observe that both too large and too small degrade the metrics of our CDNet, for instance, when $\lambda$ is set to 0.02 or 0.4. Therefore, the $\lambda$ is set to 0.1 in the experiments.

\begin{table}[]
\scriptsize
\begin{center}
\caption{The hyperparameter searching results on ZJUOUSD for $\lambda$, which is set to 0.02, 0.05, 0.1, 0.2, and 0.4.
\label{tab:lambda_ablation} }
\begin{tabular}{c|cccccc}
\Xhline{1pt}
$\lambda$ & F1 (\%)        & Acc (\%)       & Pre (\%)       & Sen (\%)       & Spe (\%)       & AUC (\%)       \\ \hline
0.02      & 93.87          & 94.18          & 94.11          & 93.73          & 98.54          & \textbf{99.40} \\
0.05      & 94.07          & 94.37          & 94.30          & 93.90          & 98.59          & 99.30          \\
0.1       & \textbf{94.13} & \textbf{94.40} & \textbf{94.43} & \textbf{93.93} & \textbf{98.59} & 99.18          \\
0.2       & 93.42          & 93.63          & 93.80          & 93.17          & 98.39          & 99.22          \\
0.4       & 91.69          & 92.07          & 92.45          & 91.47          & 98.01          & 98.75          \\ \Xhline{1pt}
\end{tabular}
\end{center}
\end{table}

\subsubsection{Analysis of Feature Distribution}
We visualize the features before the FC head of each model using UMAP \cite{mcinnes2018umap}, a technique to visualize high-dimensional features by identifying critical structures in the high-dimensional space and reserving them in the lower-dimensional embedding (e.g., 2-dimension). Compared with coarse-grained (i.e., Fig. \ref{fig:umap_ZJUOUSD} (a)), variants armed with WSLL (i.e., Fig. \ref{fig:umap_ZJUOUSD} (b-d)) have closer feature distribution among normal and ocular abnormalities. 

Moreover, we further introduce the Davies-Bouldin Score (DB Score) \cite{davies1979cluster} to measure the clustering performance, demonstrating the feature-aligned capacity of the variants of our model. All the lower-dimensional embeddings are normalized before calculating DB Score. Formally, the DB score for a cluster $c_i$ is expressed as Eq. (\ref{eq:DB_score}).

\begin{equation}
\label{eq:DB_score}
\begin{aligned}
    DB score = \max \limits_{c_i \neq c_j} \frac{Intra(c_i) + Intra(c_j)}{Inter(c_i, c_j)}
\end{aligned}
\end{equation}

\noindent where $Intra(c_i)$ represents the intra-category variance of the cluster $c_i$, and $Inter(c_i,c_j)$ represents the distance between the cluster centers of $c_i$ and $c_j$. The $Intra(c_i)$ is calculated by averaging squared deviations from the category center. Intuitively, closer and less incorrect clustering feature distribution declares finer recognition. As shown in Fig. \ref{fig:umap_ZJUOUSD}, the CDNet armed with the HCD-Loss obtains the minimum DB Score (0.3615) among all variants, depicting a closer cluster for categorization than others. Specifically, the feature distribution with the maximum DB score (i.e., 0.8772) from the baseline (i.e., Fig. \ref{fig:umap_ZJUOUSD} (a)) is more dispersed than the others. An improvement in the DB score can be observed in Fig. \ref{fig:umap_ZJUOUSD} comparing the (c) and the (d), where closer RD cases are visible and less incorrect VH cases are entangled with normal and PSS cases.

\subsubsection{Analysis of Prediction Distribution}

\begin{figure*}[ht]
  \centering
  \includegraphics[width=0.85\textwidth]{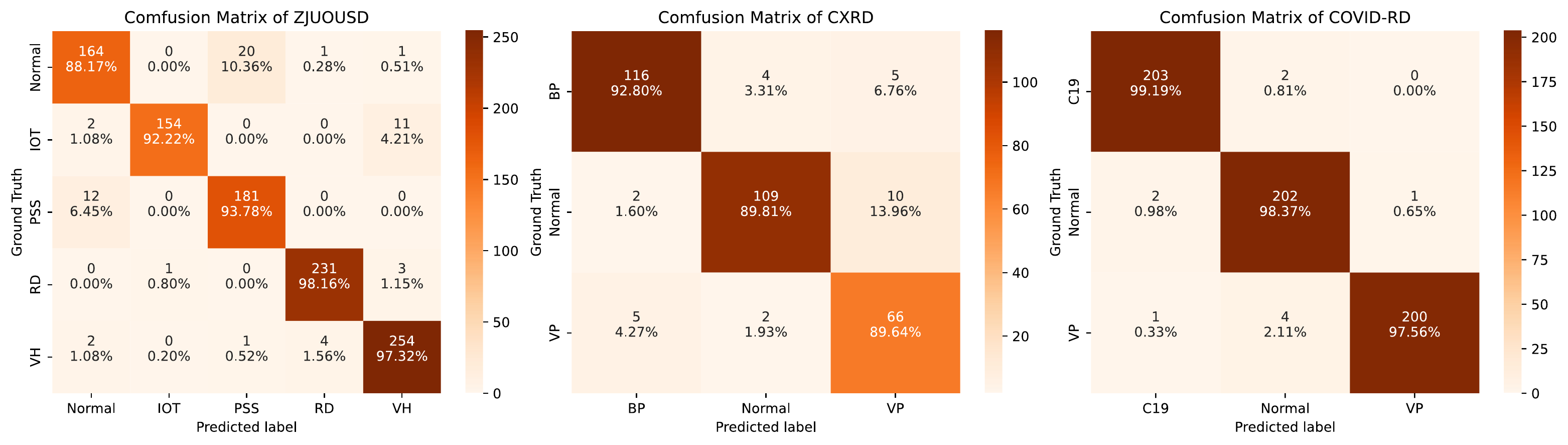}
  \caption{Three charts of confusion matrices generated by the proposed CDNet. The confusion matrices depict the distribution of predictions in three datasets. The color of confusion matrices relies on the normalized values of predictions for a better visualization, which is placed at the right of each matrix. The number of predictions is placed above the normalized values. 
  \label{fig:conf_mat}}
\end{figure*}

The distribution of prediction results is essential for assessing the efficacy of models, which is illustrated via the confusion matrix in Fig. \ref{fig:conf_mat}. The CDNet is selected to obtain the charts to display the categorization result of deep learning models. It can be implied that the CDNet can extract robust and representative features from various samples in varied modalities as most samples are on the diagonal of matrices. Moreover, these results are consistent with the feature distributions in Fig. \ref{fig:umap_ZJUOUSD}, indicating that the visible differences between IOT and PSS are incredibly subtle.

\section{Conclusion}
\label{sec:Conclusion}

In this work, the OUS recognition tasks are combined as a five-category classification task, for which we propose a novel CDNet in a feature disentanglement paradigm. The WSLL enables our CDNet to locate the lesion without relying on strong annotations while providing an explicable diagnosis decision. The CMZ strategy and the HCD-Loss collaborate for feature disentanglement. We point out that disentangling both input images and output representations can facilitate the performance of ocular abnormalities diagnosis. We demonstrate the CDNet, which significantly surpasses the previous methods in the ZJUOUSD.

Nonetheless, there are some limitations to our study. \textbf{1)} Our CDNet requires triple forward propagations (typically, one global input and two zoom-in inputs) for training, which increases the training time. \textbf{2)} 3D data, for instance, CT and MRI volumes, are not considered in our current work. \textbf{3)} The CDNet is a weakly-supervised learning method that must be trained with category-labeled data. New disease occurrence or rare diseases without large labeled images may not be classified well, while the self-supervised mechanism may be a solution to alleviate the limitations. Therefore,  we will attempt to reduce the forward times and extend the CDNet to other visual tasks in future works. Moreover, the few-shot and zero-shot settings will be explored to investigate the feature disentanglement capacity. 

\appendices

\bibliographystyle{ieeetr}
\bibliography{paper}

\begin{thebibliography}{10}

\bibitem{stannard2013radiotherapy}
C.~Stannard, W.~Sauerwein, G.~Maree, and K.~Lecuona, ``Radiotherapy for ocular
  tumours,'' {\em Eye}, vol.~27, no.~2, pp.~119--127, 2013.

\bibitem{ohno2019posterior}
K.~Ohno-Matsui and J.~B. Jonas, ``Posterior staphyloma in pathologic myopia,''
  {\em Progress in retinal and eye research}, vol.~70, pp.~99--109, 2019.

\bibitem{blindness2021vision}
G.~Blindness, ``Vision impairment, c.; vision loss expert group of the global
  burden of disease, s., trends in prevalence of blindness and distance and
  near vision impairment over 30 years: an analysis for the global burden of
  disease study,'' {\em The Lancet. Global health}, vol.~9, no.~2,
  pp.~e130--e143, 2021.

\bibitem{marques2021global}
A.~P. Marques, J.~Ramke, J.~Cairns, T.~Butt, J.~H. Zhang, D.~Muirhead,
  I.~Jones, B.~A.~A. Tong, B.~K. Swenor, H.~Faal, {\em et~al.}, ``Global
  economic productivity losses from vision impairment and blindness,'' {\em
  EClinicalMedicine}, vol.~35, p.~100852, 2021.

\bibitem{koh2020novel}
J.~E.~W. Koh, U.~Raghavendra, A.~Gudigar, O.~C. Ping, F.~Molinari, S.~Mishra,
  S.~Mathavan, R.~Raman, and U.~R. Acharya, ``A novel hybrid approach for
  automated detection of retinal detachment using ultrasound images,'' {\em
  Computers in Biology and Medicine}, vol.~120, p.~103704, 2020.

\bibitem{neupane2018imaging}
R.~Neupane, R.~Gaudana, and S.~H. Boddu, ``Imaging techniques in the diagnosis
  and management of ocular tumors: prospects and challenges,'' {\em The AAPS
  Journal}, vol.~20, no.~6, pp.~1--12, 2018.

\bibitem{blaivas2002study}
M.~Blaivas, D.~Theodoro, and P.~R. Sierzenski, ``A study of bedside ocular
  ultrasonography in the emergency department,'' {\em Academic emergency
  medicine}, vol.~9, no.~8, pp.~791--799, 2002.

\bibitem{yoonessi2010bedside}
R.~Yoonessi, A.~Hussain, and T.~B. Jang, ``Bedside ocular ultrasound for the
  detection of retinal detachment in the emergency department,'' {\em Academic
  Emergency Medicine}, vol.~17, no.~9, pp.~913--917, 2010.

\bibitem{simonyan2014very}
K.~Simonyan and A.~Zisserman, ``Very deep convolutional networks for
  large-scale image recognition,'' {\em arXiv preprint arXiv:1409.1556}, 2014.

\bibitem{ioffe2015batch}
S.~Ioffe and C.~Szegedy, ``Batch normalization: Accelerating deep network
  training by reducing internal covariate shift,'' in {\em International
  conference on machine learning}, pp.~448--456, PMLR, 2015.

\bibitem{he2016deep}
K.~He, X.~Zhang, S.~Ren, and J.~Sun, ``Deep residual learning for image
  recognition,'' in {\em Proceedings of the IEEE conference on computer vision
  and pattern recognition}, pp.~770--778, 2016.

\bibitem{huang2017densely}
G.~Huang, Z.~Liu, L.~Van Der~Maaten, and K.~Q. Weinberger, ``Densely connected
  convolutional networks,'' in {\em Proceedings of the IEEE conference on
  computer vision and pattern recognition}, pp.~4700--4708, 2017.

\bibitem{tan2019efficientnet}
M.~Tan and Q.~Le, ``Efficientnet: Rethinking model scaling for convolutional
  neural networks,'' in {\em International conference on machine learning},
  pp.~6105--6114, PMLR, 2019.

\bibitem{li2021multiscale}
J.~Li, Y.~Wang, S.~Wang, J.~Wang, J.~Liu, Q.~Jin, and L.~Sun, ``Multiscale
  attention guided network for covid-19 diagnosis using chest x-ray images,''
  {\em IEEE Journal of Biomedical and Health Informatics}, vol.~25, no.~5,
  pp.~1336--1346, 2021.

\bibitem{li2021agmb}
Y.~Li, G.~Zeng, Y.~Zhang, J.~Wang, Q.~Jin, L.~Sun, Q.~Zhang, Q.~Lian, G.~Qian,
  N.~Xia, {\em et~al.}, ``Agmb-transformer: Anatomy-guided multi-branch
  transformer network for automated evaluation of root canal therapy,'' {\em
  IEEE Journal of Biomedical and Health Informatics}, 2021.

\bibitem{xing2020zoom}
X.~Xing, Y.~Yuan, and M.~Q.-H. Meng, ``Zoom in lesions for better diagnosis:
  Attention guided deformation network for wce image classification,'' {\em
  IEEE Transactions on Medical Imaging}, vol.~39, no.~12, pp.~4047--4059, 2020.

\bibitem{hong2021disentangling}
Y.~Hong, S.~Han, K.~Choi, S.~Seo, B.~Kim, and B.~Chang, ``Disentangling label
  distribution for long-tailed visual recognition,'' in {\em Proceedings of the
  IEEE/CVF Conference on Computer Vision and Pattern Recognition},
  pp.~6626--6636, 2021.

\bibitem{zhao2022local}
S.-X. Zhao, Y.~Chen, K.-F. Yang, Y.~Luo, B.-Y. Ma, and Y.-J. Li, ``A local and
  global feature disentangled network: Toward classification of
  benign-malignant thyroid nodules from ultrasound image,'' {\em IEEE
  Transactions on Medical Imaging}, 2022.

\bibitem{feng2021two}
S.~Feng, B.~Liu, Y.~Zhang, X.~Zhang, and Y.~Li, ``Two-stream compare and
  contrast network for vertebral compression fracture diagnosis,'' {\em IEEE
  Transactions on Medical Imaging}, vol.~40, no.~9, pp.~2496--2506, 2021.

\bibitem{huang2020rectifying}
Y.-J. Huang, W.~Liu, X.~Wang, Q.~Fang, R.~Wang, Y.~Wang, H.~Chen, H.~Chen,
  D.~Meng, and L.~Wang, ``Rectifying supporting regions with mixed and active
  supervision for rib fracture recognition,'' {\em IEEE Transactions on Medical
  Imaging}, vol.~39, no.~12, pp.~3843--3854, 2020.

\bibitem{gupta2020novel}
R.~Gupta, V.~Gupta, B.~Kumar, P.~K. Singh, and A.~K. Singh, ``A novel method
  for automatic retinal detachment detection and estimation using ocular
  ultrasound image,'' {\em Multimedia Tools and applications}, vol.~79, no.~15,
  pp.~11143--11161, 2020.

\bibitem{wang2021cataract}
Y.~Wang, C.~Tang, J.~Wang, Y.~Sang, and J.~Lv, ``Cataract detection based on
  ocular b-ultrasound images by collaborative monitoring deep learning,'' {\em
  Knowledge-Based Systems}, vol.~231, p.~107442, 2021.

\bibitem{zhang2020attention}
X.~Zhang, J.~Lv, H.~Zheng, and Y.~Sang, ``Attention-based multi-model ensemble
  for automatic cataract detection in b-scan eye ultrasound images,'' in {\em
  2020 international joint conference on neural networks (IJCNN)}, pp.~1--10,
  IEEE, 2020.

\bibitem{wu2021automatic}
H.~Wu, J.~Lv, and J.~Wang, ``Automatic cataract detection with multi-task
  learning,'' in {\em 2021 International Joint Conference on Neural Networks
  (IJCNN)}, pp.~1--8, IEEE, 2021.

\bibitem{selvaraju2017grad}
R.~R. Selvaraju, M.~Cogswell, A.~Das, R.~Vedantam, D.~Parikh, and D.~Batra,
  ``Grad-cam: Visual explanations from deep networks via gradient-based
  localization,'' in {\em Proceedings of the IEEE international conference on
  computer vision}, pp.~618--626, 2017.

\bibitem{rao2021studying}
A.~Rao, J.~Park, S.~Woo, J.-Y. Lee, and O.~Aalami, ``Studying the effects of
  self-attention for medical image analysis,'' in {\em Proceedings of the
  IEEE/CVF International Conference on Computer Vision}, pp.~3416--3425, 2021.

\bibitem{wang2017zoom}
Z.~Wang, Y.~Yin, J.~Shi, W.~Fang, H.~Li, and X.~Wang, ``Zoom-in-net: Deep
  mining lesions for diabetic retinopathy detection,'' in {\em International
  Conference on Medical Image Computing and Computer-Assisted Intervention},
  pp.~267--275, Springer, 2017.

\bibitem{dong2018reinforced}
N.~Dong, M.~Kampffmeyer, X.~Liang, Z.~Wang, W.~Dai, and E.~Xing, ``Reinforced
  auto-zoom net: towards accurate and fast breast cancer segmentation in
  whole-slide images,'' in {\em Deep Learning in Medical Image Analysis and
  Multimodal Learning for Clinical Decision Support}, pp.~317--325, Springer,
  2018.

\bibitem{zhang2021ultrasound}
J.~Zhang, Q.~He, Y.~Xiao, H.~Zheng, C.~Wang, and J.~Luo, ``Ultrasound image
  reconstruction from plane wave radio-frequency data by self-supervised deep
  neural network,'' {\em Medical Image Analysis}, vol.~70, p.~102018, 2021.

\bibitem{zhou2021preservational}
H.-Y. Zhou, C.~Lu, S.~Yang, X.~Han, and Y.~Yu, ``Preservational learning
  improves self-supervised medical image models by reconstructing diverse
  contexts,'' in {\em Proceedings of the IEEE/CVF International Conference on
  Computer Vision}, pp.~3499--3509, 2021.

\bibitem{khosla2020supervised}
P.~Khosla, P.~Teterwak, C.~Wang, A.~Sarna, Y.~Tian, P.~Isola, A.~Maschinot,
  C.~Liu, and D.~Krishnan, ``Supervised contrastive learning,'' {\em Advances
  in Neural Information Processing Systems}, vol.~33, pp.~18661--18673, 2020.

\bibitem{hu2021semi}
X.~Hu, D.~Zeng, X.~Xu, and Y.~Shi, ``Semi-supervised contrastive learning for
  label-efficient medical image segmentation,'' in {\em International
  Conference on Medical Image Computing and Computer-Assisted Intervention},
  pp.~481--490, Springer, 2021.

\bibitem{chartsias2021contrastive}
A.~Chartsias, S.~Gao, A.~Mumith, J.~Oliveira, K.~Bhatia, B.~Kainz, and
  A.~Beqiri, ``Contrastive learning for view classification of
  echocardiograms,'' in {\em International Workshop on Advances in Simplifying
  Medical Ultrasound}, pp.~149--158, Springer, 2021.

\bibitem{cole2021does}
E.~Cole, X.~Yang, K.~Wilber, O.~Mac~Aodha, and S.~Belongie, ``When does
  contrastive visual representation learning work?,'' {\em arXiv preprint
  arXiv:2105.05837}, 2021.

\bibitem{pezeshki2021gradient}
M.~Pezeshki, O.~Kaba, Y.~Bengio, A.~C. Courville, D.~Precup, and G.~Lajoie,
  ``Gradient starvation: A learning proclivity in neural networks,'' {\em
  Advances in Neural Information Processing Systems}, vol.~34, 2021.

\bibitem{xu2021variational}
J.~Xu, H.~Le, M.~Huang, S.~Athar, and D.~Samaras, ``Variational feature
  disentangling for fine-grained few-shot classification,'' in {\em Proceedings
  of the IEEE/CVF International Conference on Computer Vision}, pp.~8812--8821,
  2021.

\bibitem{pei2021disentangle}
C.~Pei, F.~Wu, L.~Huang, and X.~Zhuang, ``Disentangle domain features for
  cross-modality cardiac image segmentation,'' {\em Medical Image Analysis},
  vol.~71, p.~102078, 2021.

\bibitem{cheng2021multimodal}
J.~Cheng, M.~Gao, J.~Liu, H.~Yue, H.~Kuang, J.~Liu, and J.~Wang, ``Multimodal
  disentangled variational autoencoder with game theoretic interpretability for
  glioma grading,'' {\em IEEE Journal of Biomedical and Health Informatics},
  2021.

\bibitem{bau2019seeing}
D.~Bau, J.-Y. Zhu, J.~Wulff, W.~Peebles, H.~Strobelt, B.~Zhou, and A.~Torralba,
  ``Seeing what a gan cannot generate,'' in {\em Proceedings of the IEEE/CVF
  International Conference on Computer Vision}, pp.~4502--4511, 2019.

\bibitem{schroff2015facenet}
F.~Schroff, D.~Kalenichenko, and J.~Philbin, ``Facenet: A unified embedding for
  face recognition and clustering,'' in {\em Proceedings of the IEEE conference
  on computer vision and pattern recognition}, pp.~815--823, 2015.

\bibitem{wang2018deep}
Y.~Wang, Z.~Deng, X.~Hu, L.~Zhu, X.~Yang, X.~Xu, P.-A. Heng, and D.~Ni, ``Deep
  attentional features for prostate segmentation in ultrasound,'' in {\em
  International Conference on Medical Image Computing and Computer-Assisted
  Intervention}, pp.~523--530, Springer, 2018.

\bibitem{han2020learning}
Z.~Han, Z.~Fu, and J.~Yang, ``Learning the redundancy-free features for
  generalized zero-shot object recognition,'' in {\em Proceedings of the
  IEEE/CVF Conference on Computer Vision and Pattern Recognition},
  pp.~12865--12874, 2020.

\bibitem{zhang2021knowledge}
S.~Zhang, R.~Du, D.~Chang, Z.~Ma, and J.~Guo, ``Knowledge transfer based
  fine-grained visual classification,'' in {\em 2021 IEEE International
  Conference on Multimedia and Expo (ICME)}, pp.~1--6, IEEE, 2021.

\bibitem{peng2022crafting}
X.~Peng, K.~Wang, Z.~Zhu, and Y.~You, ``Crafting better contrastive views for
  siamese representation learning,'' {\em arXiv preprint arXiv:2202.03278},
  2022.

\bibitem{huang2019ccnet}
Z.~Huang, X.~Wang, L.~Huang, C.~Huang, Y.~Wei, and W.~Liu, ``Ccnet: Criss-cross
  attention for semantic segmentation,'' in {\em Proceedings of the IEEE/CVF
  International Conference on Computer Vision}, pp.~603--612, 2019.

\bibitem{hou2019cross}
R.~Hou, H.~Chang, B.~Ma, S.~Shan, and X.~Chen, ``Cross attention network for
  few-shot classification,'' {\em Advances in Neural Information Processing
  Systems}, vol.~32, 2019.

\bibitem{zbontar2021barlow}
J.~Zbontar, L.~Jing, I.~Misra, Y.~LeCun, and S.~Deny, ``Barlow twins:
  Self-supervised learning via redundancy reduction,'' in {\em International
  Conference on Machine Learning}, pp.~12310--12320, PMLR, 2021.

\bibitem{chen2021crossvit}
C.-F.~R. Chen, Q.~Fan, and R.~Panda, ``Crossvit: Cross-attention multi-scale
  vision transformer for image classification,'' in {\em Proceedings of the
  IEEE/CVF International Conference on Computer Vision}, pp.~357--366, 2021.

\bibitem{kermany2018identifying}
D.~S. Kermany, M.~Goldbaum, W.~Cai, C.~C. Valentim, H.~Liang, S.~L. Baxter,
  A.~McKeown, G.~Yang, X.~Wu, F.~Yan, {\em et~al.}, ``Identifying medical
  diagnoses and treatable diseases by image-based deep learning,'' {\em Cell},
  vol.~172, no.~5, pp.~1122--1131, 2018.

\bibitem{rahman2021exploring}
T.~Rahman, A.~Khandakar, Y.~Qiblawey, A.~Tahir, S.~Kiranyaz, S.~B.~A. Kashem,
  M.~T. Islam, S.~Al~Maadeed, S.~M. Zughaier, M.~S. Khan, {\em et~al.},
  ``Exploring the effect of image enhancement techniques on covid-19 detection
  using chest x-ray images,'' {\em Computers in biology and medicine},
  vol.~132, p.~104319, 2021.

\bibitem{chowdhury2020can}
M.~E. Chowdhury, T.~Rahman, A.~Khandakar, R.~Mazhar, M.~A. Kadir, Z.~B. Mahbub,
  K.~R. Islam, M.~S. Khan, A.~Iqbal, N.~Al~Emadi, {\em et~al.}, ``Can ai help
  in screening viral and covid-19 pneumonia?,'' {\em IEEE Access}, vol.~8,
  pp.~132665--132676, 2020.

\bibitem{lv2021cascade}
D.~Lv, Y.~Wang, S.~Wang, Q.~Zhang, W.~Qi, Y.~Li, and L.~Sun, ``A cascade-seme
  network for covid-19 detection in chest x-ray images,'' {\em Medical
  Physics}, vol.~48, no.~5, pp.~2337--2353, 2021.

\bibitem{rundo2021advanced}
F.~Rundo, A.~Genovese, R.~Leotta, F.~Scotti, V.~Piuri, and S.~Battiato,
  ``Advanced 3d deep non-local embedded system for self-augmented x-ray-based
  covid-19 assessment,'' in {\em Proceedings of the IEEE/CVF International
  Conference on Computer Vision}, pp.~423--432, 2021.

\bibitem{info11020125}
A.~Buslaev, V.~I. Iglovikov, E.~Khvedchenya, A.~Parinov, M.~Druzhinin, and
  A.~A. Kalinin, ``Albumentations: Fast and flexible image augmentations,''
  {\em Information}, vol.~11, no.~2, 2020.

\bibitem{mcinnes2018umap}
L.~McInnes, J.~Healy, and J.~Melville, ``Umap: Uniform manifold approximation
  and projection for dimension reduction,'' {\em arXiv preprint
  arXiv:1802.03426}, 2018.

\bibitem{davies1979cluster}
D.~L. Davies and D.~W. Bouldin, ``A cluster separation measure,'' {\em IEEE
  transactions on pattern analysis and machine intelligence}, no.~2,
  pp.~224--227, 1979.

\end{thebibliography}

\end{document}